\documentclass[sigconf]{acmart}
\AtBeginDocument{%
  }

\setcopyright{acmlicensed}
\copyrightyear{2018}
\acmYear{2018}
\acmDOI{XXXXXXX.XXXXXXX}
\acmConference[Conference acronym 'XX]{Make sure to enter the correct
  conference title from your rights confirmation email}{June 03--05,
  2018}{Woodstock, NY}
\acmISBN{978-1-4503-XXXX-X/2018/06}

\usepackage{booktabs,makecell,graphicx}
\usepackage{xcolor,colortbl}
\usepackage{xcolor}
\usepackage{booktabs,tabularx,array,colortbl}
\definecolor{AtlasAnswer}{HTML}{B86729}
\definecolor{AtlasAnswerBg}{HTML}{FFF2E3}
\definecolor{AtlasTrajectory}{HTML}{7A56A5}
\definecolor{AtlasTrajectoryBg}{HTML}{F1EBF8}
\definecolor{AtlasEfficiency}{HTML}{2B78AA}
\definecolor{AtlasEfficiencyBg}{HTML}{EAF3FA}
\definecolor{AtlasGreen}{HTML}{238875}
\definecolor{AtlasGroupBg}{HTML}{EFF6F2}
\definecolor{AtlasInk}{HTML}{26363C}
\definecolor{AtlasMuted}{HTML}{63737B}
\definecolor{AtlasRule}{HTML}{9BAAAF}
\definecolor{AtlasStripe}{HTML}{F7F9FA}
\definecolor{AtlasToolBlue}{HTML}{2878A6}
\definecolor{AtlasRevisionPurple}{HTML}{8056B3}
\definecolor{AtlasDiversityGreen}{HTML}{278975}
\newcommand{\AtlasHead}[1]{\makecell[c]{#1}}
\newlength{\AtlasModelWidth}
\newcommand{\AtlasModel}[2]{%
  \parbox[c]{\AtlasModelWidth}{\raggedright\strut\mbox{#1}\\[-0.5pt]{\fontsize{6.5}{7.5}\selectfont\color{AtlasMuted}(#2)\strut}}}
\newcommand{\AtlasFamily}[1]{%
  \rowcolor{AtlasGroupBg}%
  \multicolumn{13}{l}{\strut\textcolor{AtlasGreen}{\bfseries #1}}\\}
\usepackage{pifont} % 其余沿用已有的 booktabs、xcolor、array

\DeclareRobustCommand{\BenchYes}{%
    \textcolor{AtlasGreen}{\ding{51}}}
\DeclareRobustCommand{\BenchPartial}{%
    \textcolor{AtlasTrajectory}{\ensuremath{\triangle}}}
\DeclareRobustCommand{\BenchNo}{%
    \textcolor{AtlasMuted}{\ding{55}}}

\usepackage{array}
\usepackage{booktabs}
\usepackage{tabularx}
\usepackage{array}
\usepackage{makecell}
\renewcommand\footnotetextcopyrightpermission[1]{} % 隐藏左下角的版权信息/会议信息块
\begin{document}

\title{ATLAS-Bench: An Agentic Tool-Use Benchmark for Local Area Search}

\author{Keyun Zhang}
\authornote{Equal contribution.}
\affiliation{\institution{AMAP, Alibaba Group}\country{Beijing, China}}
\email{zhangkeyun.zky@alibaba-inc.com}

\author{Dapeng Yin}
\authornotemark[1]
\affiliation{\institution{AMAP, Alibaba Group}\country{Beijing, China}}
\email{yindapeng.ydp@alibaba-inc.com}

\author{Fanfu Wang}
\authornotemark[1]
\affiliation{\institution{AMAP, Alibaba Group}\country{Beijing, China}}
\email{wangfanfu.wff@alibaba-inc.com}

\author{Shuang Lang}
\affiliation{\institution{AMAP, Alibaba Group}\country{Beijing, China}}
\email{langshuang.ls@alibaba-inc.com}

\author{Zhenyu Zhu}
\affiliation{\institution{AMAP, Alibaba Group}\country{Beijing, China}}
\email{hanyue.zzy@alibaba-inc.com}

\author{Yulin Huang}
\affiliation{\institution{AMAP, Alibaba Group}\country{Beijing, China}}
\email{huangyulin.hyl@alibaba-inc.com}

\author{Chong Yang}
\authornote{Corresponding author.}
\affiliation{\institution{AMAP, Alibaba Group}\country{Beijing, China}}
\email{hanli.yc@alibaba-inc.com}

\author{\texorpdfstring{{Pengjie Wang}}{Pengjie Wang}}
\affiliation{\institution{AMAP, Alibaba Group}\country{Beijing, China}}
\email{pengjie.wpj@alibaba-inc.com}

% Explain the underlined name without adding another superscript symbol.
\makeatletter
\g@addto@macro\@authornotes{%
  \begingroup
  \renewcommand{\thefootnote}{}%
  \footnotetext{{Pengjie Wang}: Project leader.}%
  \endgroup}
\makeatother

\renewcommand{\shortauthors}{Zhang et al.}

\begin{abstract}
Reliable local search is essential for LLM agents to support everyday decisions and longer plans by finding places that meet users’ needs at given times and locations. Yet existing benchmarks provide limited coverage of whether search results satisfy mandatory requirements, account for preferences, and ground factual and experiential judgments in place attributes and reviews.
We introduce ATLAS-Bench, a benchmark of 1,200 queries across three regions, expanded from real search events using place attributes, reviews, and existing user profiles from the platform. Each query distinguishes mandatory requirements from preferences and factual conditions from subjective judgments, with supporting evidence and criteria for checking each requirement. Queries are validated through iterative assertion checks and cross-model review.
We also introduce ATLASBox, a standardized evaluation environment with search tools and a fixed point-of-interest (POI) knowledge base. Given a query, timestamp, and location, an agent searches place attributes and reviews to recommend evidence-supported options or report that no place meets all mandatory requirements. We evaluate answer quality, trajectory quality, and execution cost.
% Among the 11 evaluated models,
Across 14 configurations of 11 models,
GPT-6-Luna achieves the highest constraint satisfaction rate among returned recommendations (71.5\%) and factual accuracy (95.6\%). However, it returns at least one acceptable
recommendation for fewer satisfiable queries than GLM-5.3 (67.3\%
versus 85.3\%). These results highlight the challenge of satisfying individual
requirements while providing usable recommendations across queries. 

\end{abstract}






\maketitle

\section{Introduction}

LLM-based search agents handle complex information needs on the Web by retrieving iteratively, calling Web APIs and tools, and combining evidence from multiple online sources~\cite{li2025search,li2026webthinker,song2025r1searcherincentivizingsearchcapability}.
Unlike conventional Web retrieval, they must interpret user intent and decide whether the evidence suffices for an answer, which matters most for requests that combine multiple constraints and preferences~\cite{malaviya-etal-2023-quest,weller-etal-2025-followir,oh2024instructirbenchmarkinstructionfollowing,su2025brightrealisticchallengingbenchmark}. Evaluating such agents requires tasks grounded in real user needs and their context, together with criteria for judging whether an answer meets them.

\begin{figure}[!t]
  \centering
  \includegraphics[width=\linewidth]{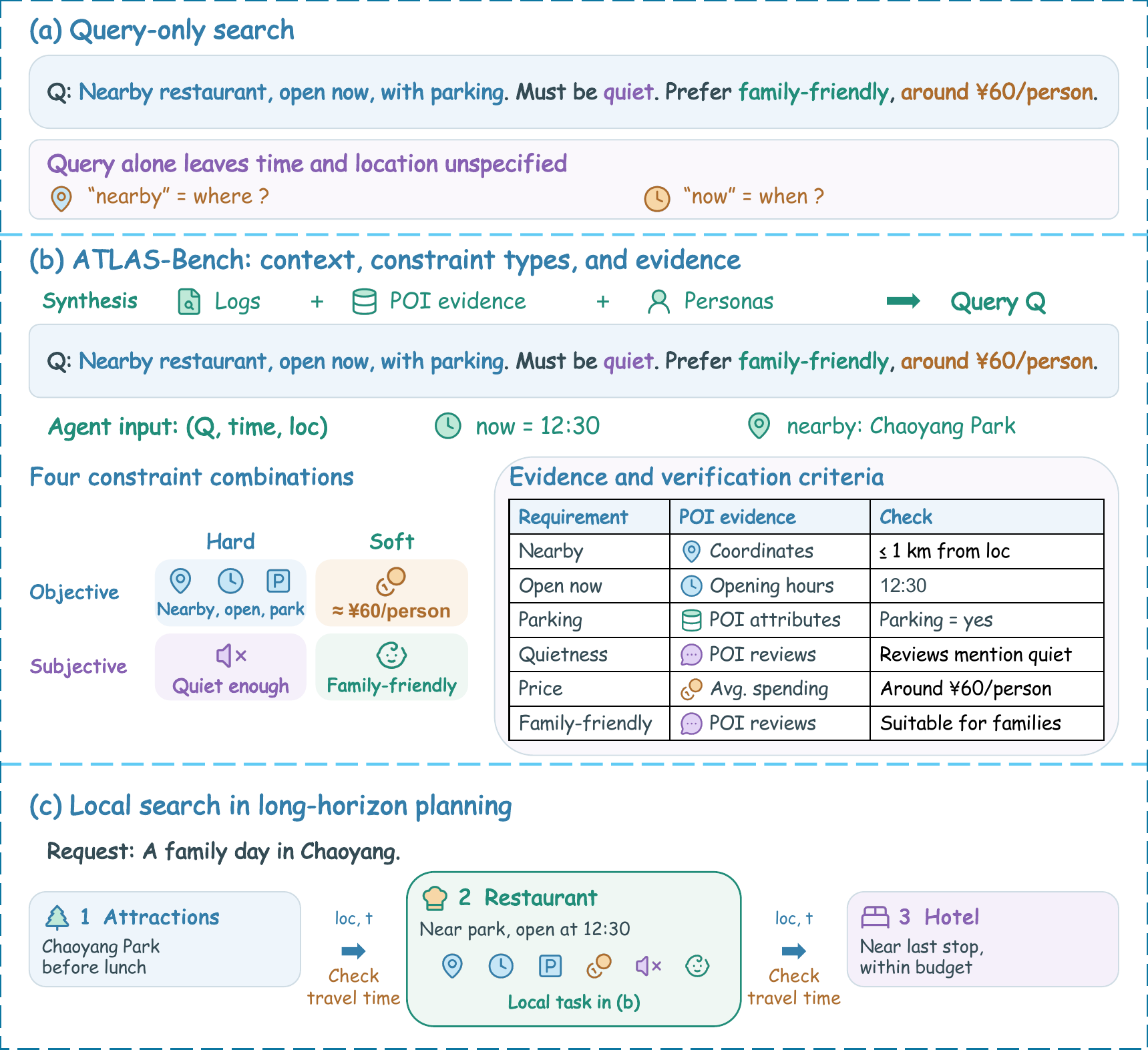}
  \caption{
Overview of ATLAS-Bench. (a) Query-only search lacks context
and treats all constraints alike. (b) ATLAS-Bench adds context,
constraint types, and evidence. (c) Local search within a longer plan.}
  \Description{An overview comparing traditional search and agentic local search with multi-faceted constraints.}
  \label{fig:1}
  \vspace{-14pt}
\end{figure}

Local area search on navigation Web platforms offers such a setting~\cite{10.1145/3589335.3648334,10.1145/3711896.3737196}.
Consider the query \textit{Nearby restaurant, open now, with parking. Must be quiet. Prefer family-friendly, around \textyen 60/person.}
Given the time and location needed to resolve ``nearby''
and ``open now'' (Figure~\ref{fig:1}(a)), an agent must search candidate restaurants, inspect their attributes and user-generated reviews, and recommend options with supporting evidence or report that none qualifies.
The requirements differ in two independent ways: hard constraints must hold while soft preferences allow trade-offs, and objective ones are checked against attributes while subjective ones are judged from reviews.
An acceptable recommendation must support every hard constraint with evidence, such as reviews for the subjective constraint of quietness, and disclose any unmet soft preference, such as the price target (Figure~\ref{fig:1}(b)).
Such a request is often one stage of a longer plan, such as lunch during a family day out, making reliable local search a building block of long-horizon planning (Figure~\ref{fig:1}(c)).

Existing benchmarks evaluate several aspects of this setting
(Table~\ref{tab:benchmark-comparison}):
MapEval~\cite{3780338.3780864} and
LocalSearchBench~\cite{10.1145/3770855.3817466} use manually constructed
queries for geospatial reasoning and multi-hop merchant search, whereas
MapSatisfyBench~\cite{bai2026mapsatisfybenchbenchmarkingsatisfactionawaremap}
and MobilityBench~\cite{10.1145/3770855.3817498} ground tasks in real
map or mobility logs.
However, none establishes a systematic constraint taxonomy with
evidence-grounded verification to handle heterogeneous user needs.
In the restaurant example, recommending a noisy restaurant and
recommending a quiet one at \textyen 80/person with the price disclosed
each meet five of six requirements, yet only the latter is acceptable.
Distinguishing such cases requires checking each requirement according
to its type and evidence.

To address this gap, we introduce \textbf{ATLAS-Bench}, an
\textbf{A}gentic \textbf{T}ool-Use Benchmark for \textbf{L}ocal
\textbf{A}rea \textbf{S}earch comprising 1,200 queries across three
regions with contrasting POI densities: Chaoyang (Beijing), Huangpu
(Shanghai), and Dabancheng (Urumqi).
Each query preserves a logged search intent while incorporating
persona-guided requirements grounded in the attributes and reviews of an
engaged reference POI; however, any POI satisfying the stated constraints counts as a valid answer, and all preferences are explicitly expressed so evaluation requires only the query, timestamp, and location.
A recursive self-improving (RSI) pipeline validates queries via
assertion checks and cross-model review, converting detected failures
into new assertions for subsequent batches and yielding fine-grained
annotations of constraint hardness, subjectivity, relations, and
verification evidence.

\begin{table}[htbp]
    \centering
    \caption{Comparison of local search and mobility benchmarks.
    \BenchYes: supported;
    \BenchPartial: partial or limited;
    \BenchNo: not supported.}
    \label{tab:benchmark-comparison}

    \begingroup
    \footnotesize
    \setlength{\tabcolsep}{2pt}
    \renewcommand{\arraystretch}{1.18}
    \arrayrulecolor{AtlasRule}
    \color{AtlasInk}

    \resizebox{\linewidth}{!}{%
    \begin{tabular}{
        @{}lcccc
        >{\columncolor{AtlasTrajectory!8}}c@{}}
        \toprule
        \textbf{Dimension}
        & \textbf{MapEval}
        & \textbf{LocalSearch}
        & \textbf{Mobility}
        & \textbf{MapSatisfy}
        & \textcolor{AtlasTrajectory}{\textbf{ATLAS}} \\
        \midrule

        Query source
        & Manual & Manual & Trajectory
        & \makecell{Map logs}
        & Log-grounded \\

        Reference target
        & Manual & Manual & Destination
        & \makecell{Decision\\rubrics}
        & Clicked POI \\

        \midrule
        Context
        & \BenchNo & \BenchPartial & \BenchYes
        & \BenchYes & \BenchYes \\

        Constraint taxonomy
        & \BenchNo & \BenchPartial & \BenchNo
        & \BenchPartial & \BenchYes \\

        Constraint conflicts
        & \BenchNo & \BenchNo & \BenchNo
        &  \BenchNo& \BenchYes \\

        Implicit needs
        & \BenchNo & \BenchPartial & \BenchNo
        & \BenchYes & \BenchYes \\

        \midrule
        Quality control
        & Manual & Manual & \makecell{Static\\rules}
        & \makecell{LLM judges}
        & \makecell{Rules\\+ Agent} \\
        \bottomrule
    \end{tabular}%
    }
    \endgroup
    \end{table}
Building on ATLAS-Bench, we develop \textbf{ATLASBox}, a reproducible
evaluation sandbox that avoids temporal drift in live map services by
serving a fixed POI knowledge base through standardized tools for POI
retrieval, nearby search, and review inspection.
ATLASBox logs all tool interactions to evaluate final answers and
trajectories across constraint satisfaction, factual accuracy,
trajectory quality, and execution cost. Our contributions are threefold:
\begin{itemize}
    \item \textbf{Taxonomized Benchmark and Reproducible Sandbox.}
    We introduce \textbf{ATLAS-Bench} (1,200 log-grounded queries refined via recursive self-improvement) and the \textbf{ATLASBox} sandbox, jointly modeling constraint hardness, subjectivity, and conflicts with traceable POI evidence for rigorous agent evaluation.

    \item \textbf{Comprehensive Benchmarking and Trade-off Analysis.}
    Evaluating 14 model configurations, we reveal a critical trade-off between task coverage and constraint satisfaction across exploratory and conservative search strategies.

    % \item \textbf{Bridging Local Search and Long-Horizon Planning.}
    % We show that while single-pass planning yields brief, unactionable results, subquery decomposition produces detailed itineraries but suffers from severe subquery errors, underscoring the criticality of atomic local search.
    %     \item \textbf{In-Depth Analysis of Challenges and Planning Impact.}
    % Analyzing regional POI density, constraint types, and long-horizon itineraries, we show that dense candidates and subjective verification remain major bottlenecks, underscoring the criticality of reliable atomic local search for complex planning.
    \item \textbf{Local Search Challenges.}
Analyzing regional settings, constraint types, and a long-horizon
case, we highlight challenges in filtering dense candidate spaces
and verifying complex, subjective requirements, underscoring
the importance of reliable local search for long-horizon planning.

\end{itemize}

\section{Related Work}

\subsection{Search Agents and Their Evaluation}
Web search agents interleave retrieval with reasoning, as in
Search-o1 and WebThinker~\cite{li2025search,li2026webthinker},
and combine task decomposition with online evidence
synthesis~\cite{zheng2025deepresearcherscalingdeepresearch,
chen2025mindsearch,wu2026webdancer}, while
Search-R1 and R1-Searcher optimize search decisions via
reinforcement learning~\cite{jin2025search,
song2025r1searcherincentivizingsearchcapability}.

Existing benchmarks evaluate related capabilities:
QUEST tests entity-seeking queries with implicit set
operations~\cite{malaviya-etal-2023-quest},
InstructIR and FollowIR assess instruction-following
retrieval~\cite{oh2024instructirbenchmarkinstructionfollowing,
weller-etal-2025-followir}, and BRIGHT targets reasoning-intensive
retrieval~\cite{su2025brightrealisticchallengingbenchmark}.
HotpotQA~\cite{yang-etal-2018-hotpotqa},
MuSiQue~\cite{trivedi-etal-2022-musique}, and
BrowseComp~\cite{wei2025browsecompsimplechallengingbenchmark}
evaluate multi-hop QA and persistent web browsing.
While these tasks target document relevance or factual QA,
ATLAS-Bench evaluates spatiotemporal local search, requiring agents to
verify heterogeneous constraints against POI attributes and user
reviews.

% \vspace{-4pt}
\subsection{Geospatial and Local Search Benchmarks}
MapEval and MapQA evaluate geospatial
reasoning~\cite{3780338.3780864,10.1145/3748636.3764174},
while LocalEval tests local-service knowledge and
user--service interactions~\cite{10.1145/3711896.3737196}.
LocalSearchBench evaluates multi-hop merchant and web
search~\cite{10.1145/3770855.3817466}, and
MapSatisfyBench focuses on inferring unspoken user needs
from behavioral histories and multi-turn clarification~\cite{bai2026mapsatisfybenchbenchmarkingsatisfactionawaremap}.

However, these benchmarks do not jointly model constraint
hardness, subjectivity, and conflicts, nor do they verify
experiential requirements against POI reviews.
ATLAS-Bench annotates both hard/soft and objective/subjective
dimensions, linking each requirement to POI attributes or
reviews under spatiotemporal context to evaluate evidence-backed
recommendations and preference trade-offs.

% \vspace{-4pt}
\subsection{Long-Horizon Planning and Executable Constraint Evaluation}
TravelPlanner checks itinerary constraints against travel
records~\cite{3692070.3694316}, while ChinaTravel evaluates
compositional constraints and soft preferences through an
executable domain-specific language over activity
attributes~\cite{shao2026chinatravel}.
MobilityBench evaluates route planning using real mobility
queries and a frozen API replay
environment~\cite{10.1145/3770855.3817498}.

TravelPlanner and ChinaTravel mainly use structured attributes and programmatic checks to verify constraints.
ATLAS-Bench also evaluates subjective requirements supported
by review evidence, such as quietness and atmosphere.
It focuses on contextual place selection as a constituent
capability of longer plans, complementing itinerary-level
feasibility evaluation.

\section{Construction of ATLAS}
\label{sec:atlas_construction}

Figure~\ref{fig:construction} outlines the construction pipeline of ATLAS. 
This section describes the task formulation (\S\ref{sec:formulation}), 
POI knowledge base construction (\S\ref{sec:kb}), 
the recursive self-improving data synthesis pipeline (\S\ref{sec:synthesis}), 
dataset statistics and comparative analysis (\S\ref{sec:stats}), 
and the integration of the ATLASBox evaluation sandbox (\S\ref{sec:sandbox}).

\vspace{-2pt}
\subsection{Problem Statement}
\label{sec:formulation}
\subsubsection{Formulation of ATLAS}
% 子约束阐述

ATLAS-Bench focuses on agentic local area search characterized by incomplete information and multi-faceted constraints across temporal, spatial, and user-centric dimensions. Unlike plain retrieval or routing tasks, agents must infer implicit intents, invoke tools to gather evidence, and arbitrate between competing constraints.

Local search also forms a constituent decision problem in longer planning workflows. For example, \textit{arranging dinner after a movie} translates the cinema location, expected finishing time, and subsequent activities into spatial, opening-hour, and dining-duration constraints. ATLAS-Bench abstracts this step as multi-constraint POI search conditioned on a query and its context, evaluating whether an agent retrieves suitable candidates supported by evidence. These candidates can inform subsequent planning decisions, while complete plans additionally require coordination of cross-step dependencies, cumulative budgets, and schedule feasibility. Performance on ATLAS-Bench therefore characterizes a component capability rather than establishing end-to-end planning competence.
To support this task, ATLAS-Bench follows three key design principles:

\textbf{Behavioral Grounding}: Each instance originates from a logged search event, retaining its
query, timestamp, privacy-processed user coordinates, and associated POI interactions. POIs associated with navigation initiation or detail-page viewing longer than five seconds serve as candidate reference targets rather than unique ground-truth answers.

% \textbf{Verifiability}: Every synthesized constraint is tied to observable evidence such as structured POI attributes or verified reviews. We adopt a target-conditioned inverse synthesis pipeline, extracting authentic evidence first before generating natural language constraints, which grounds query synthesis in traceable evidence and supports subsequent verification.
\textbf{Verifiability}: Every synthesized constraint is grounded in
observable evidence, such as structured POI attributes or verified
reviews. Our target-conditioned inverse synthesis pipeline extracts
this evidence before generating natural-language constraints, so each
query remains traceable and verifiable.

\textbf{Systematic Coverage}: Constraints are characterized by subjectivity (objective versus subjective) and hardness (hard versus soft), together with relations among requirements, to support sampling and diagnosis.
\begin{figure}[htbp]
  \centering
  \includegraphics[width=\linewidth]{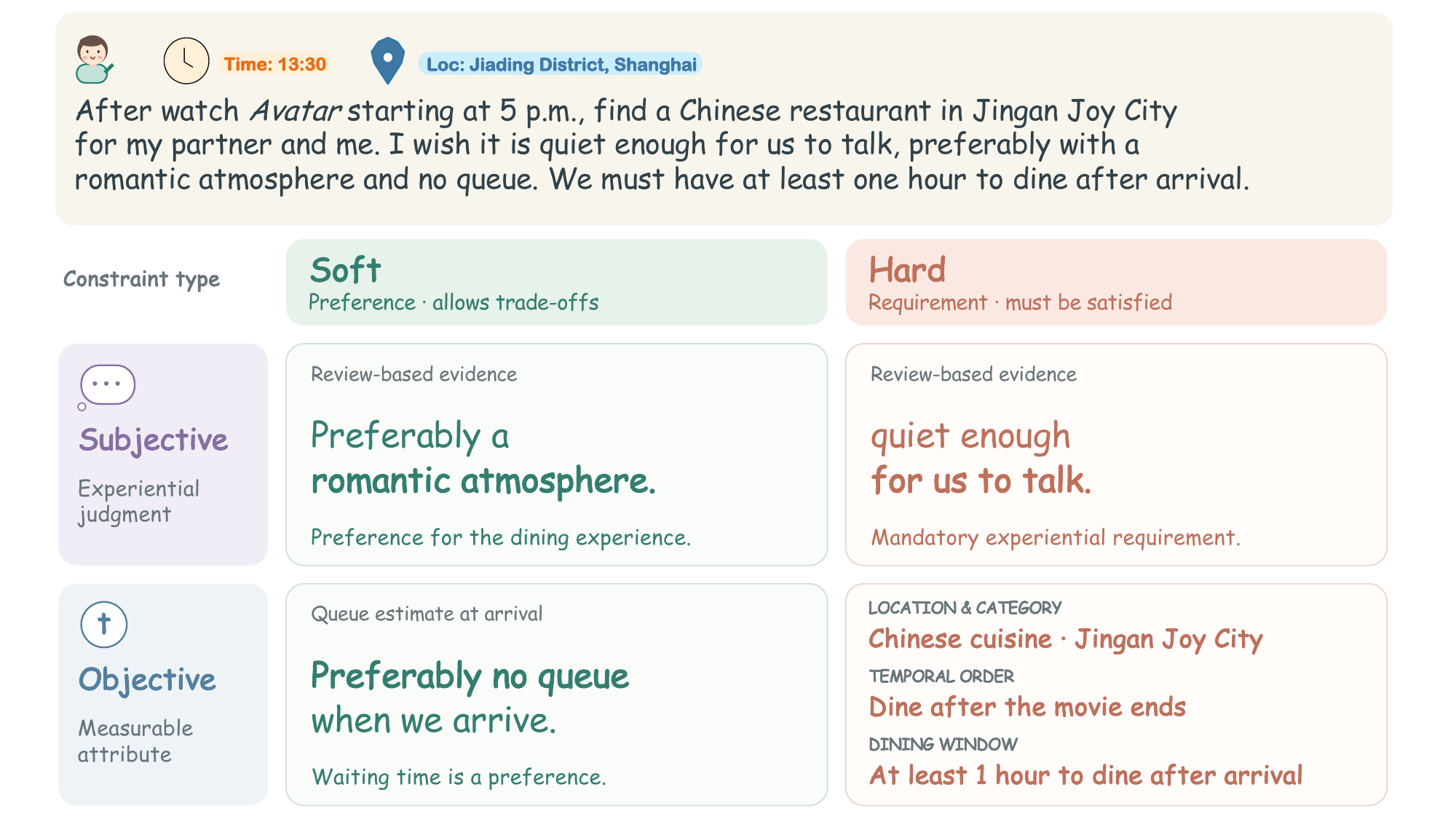}
  \caption{Four constraint combinations for a shared
  restaurant-search intent. }
  \Description{A two-by-two matrix illustrating subjective
  and objective constraints, each expressed as soft
  preferences or hard requirements.}
  \label{fig:constraint-taxonomy}
  \vspace{-12pt}
\end{figure}
\subsubsection{Constraint Taxonomy}
\label{sec:Constraint_Taxonomy}
ATLAS-Bench organizes requirements into atomic constraints
and their relations. Atomic constraints have two independent
dimensions: \textit{subjectivity}, distinguishing factual
conditions from experiential judgments, and \textit{hardness},
distinguishing mandatory requirements from preferences that
allow trade-offs. 
Figure~\ref{fig:constraint-taxonomy} illustrates the four
combinations of subjectivity and hardness within a single
restaurant-search request.
Each constraint is paired with an evidence source and a check
predicate. Constraints may be compatible, trade off against each
other, conflict as hard requirements, or depend on one another. Hard conflicts make a query unsatisfiable by design, in which case the
expected behavior is to report that no candidate meets all hard
constraints.
Task POIs serve as spatial anchors, excluded candidates, or candidate
reference targets; reference-target names never appear in the query,
avoiding answer leakage.

\subsection{POI Knowledge Base Construction}
\label{sec:kb}

\begin{figure*}[t]
  \centering
  \includegraphics[width=\textwidth]{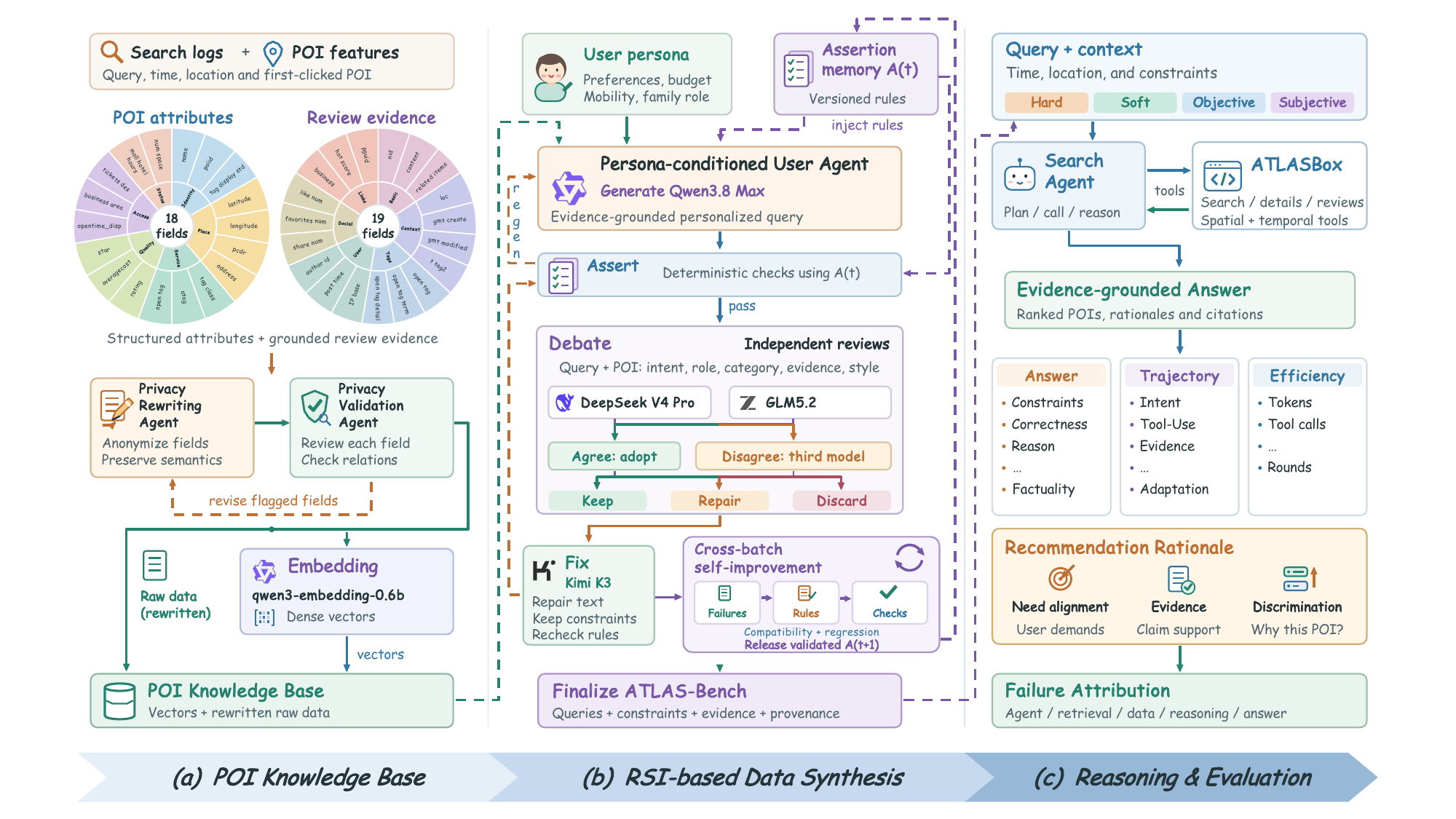}
  \caption{Overview of ATLAS.
\textbf{(a)} Search logs are linked to POI attributes and reviews,
privacy-rewritten and validated, and stored as vectors and raw records.
\textbf{(b)} Persona-aware queries pass through assertions, cross-model
debate, repair, and finalization; failures update the assertion memory
from $\mathcal{A}^{(t)}$ to $\mathcal{A}^{(t+1)}$.
\textbf{(c)} Agents solve the tasks in ATLASBox and are evaluated on answer
quality, trajectory quality, and efficiency.}

  \label{fig:construction}
\end{figure*}
The source data comprise search logs from a major commercial navigation
platform and online POI snapshots of the three target regions.
Each search event records the query, timestamp, user coordinates, and
subsequent POI interactions, including navigation initiation and
detail-page viewing duration.
% The POI knowledge base covers all active POIs in each region.
The POI knowledge base indexes all filtered online POIs across each region, not just those appearing in the search logs.
For each POI, we collect structured attributes (category, address,
rating, average spending, opening hours, facilities, and operating
status) and link its reviews and place-related notes.
Attributes and reviews are stored separately but linked by POI
identifiers, with dedicated tools for review retrieval.

User profiles originate from existing platform records and,
after privacy processing, are used only for query synthesis.
The generation agent adopts the supplied persona to formulate
requests reflecting its preferences and needs.
For example, a persona traveling by car may motivate a parking
requirement in the generated query.
At evaluation time, the search agent receives only the query,
timestamp, and user location, without separate profile fields.
Any profile-derived requirement used for scoring must therefore
be recoverable from these visible inputs, rather than imposed
solely through hidden profile attributes.

For query synthesis, we retain search events with the interactions.
The associated POI serves as a candidate reference target, whose
compatibility with the expanded query is subsequently checked against
POI attributes and textual evidence.
Other POIs in the knowledge base that satisfy the query constraints
remain valid answers.

% Before use, a Privacy Rewriting Agent transforms potentially
% identifying fields in query contexts, user profiles, and POI
% records, followed by field-level review by a Privacy Validation
% Agent.
Before use, a Privacy Rewriting Agent transforms
identifying fields in query contexts, user profiles, and POI
records, followed by review by a Privacy Validation Agent.
User profiles retain abstract attributes relevant to search,
such as spending ranges, travel modes, and family roles.
Identifiers, contact information, and address details are rewritten
or replaced, and user coordinates are perturbed,
with checks for preservation of search intent, spatiotemporal
relations, and POI characteristics. Anonymized data use for this benchmark was authorized by the data provider.
The processed textual content is encoded using
Qwen3-Embedding-0.6B~\cite{qwen3embedding}.
The knowledge base stores both embeddings and rewritten records
to support retrieval and evidence inspection, as illustrated
in Figure~\ref{fig:construction}(a).

\subsection{RSI-based Data Synthesis}
\label{sec:synthesis}

ATLAS-Bench synthesizes queries with a recursive self-improving (RSI) pipeline comprising Prepare, Generate, Assert, Debate, Fix, and Finalize (Figure~\ref{fig:construction}(b)). Self-improvement accumulates checking rules in an external assertion memory, without updating model parameters.

\subsubsection{Evidence-Grounded Generation}
\label{sec:rsi_prep_gen}

Prepare groups candidate records by POI category and assigns expression
styles, anchor types, and target difficulty levels.
Qwen3.8-Max~\cite{qwen38} serves as the main agent: it generates
queries, arbitrates review disagreements
(Section~\ref{sec:rsi_assertions}), and maintains the assertion memory
(Section~\ref{sec:rsi_evolution}).
In Generate, it acts as a persona-conditioned user agent
(Section~\ref{sec:kb}) that preserves the logged query's core intent
and spatiotemporal context, retrieves attributes or reviews of the
candidate reference POI, and adds only evidence-supported constraints,
which are synthesized rather than observed in the original query.
Each query is annotated with subjectivity, hardness, evidence sources,
and expected predicates.

\subsubsection{Assertion Checking and Cross-Model Review}
\label{sec:rsi_assertions}

Assert checks each query against the current assertion memory, covering POI roles, target-name leakage, geographic validity, retrieval intent, category compatibility, text integrity, target status, and support for annotated conflicts; deterministic conditions are checked directly, and semantic ones follow the criteria in memory. Passing queries enter Debate, where DeepSeek-V4-Pro~\cite{deepseekai2026deepseekv4} and GLM-5.2~\cite{glm5team2026glm5vibecodingagentic} independently assess the query, reference target, and evidence for consistency and naturalness. Joint acceptance retains a query, joint rejection routes it to repair or removal, and disagreement is arbitrated by the main agent in a separate context. Fix uses Kimi-K3~\cite{kimiteam2026kimik3openfrontier} to repair wording while preserving constraints, evidence links, and target semantics; substantive defects require regeneration or removal. Repaired and regenerated queries restart from Assert until a retry limit. Finalize derives constraint labels, removes duplicates, and retains context, reference targets, evidence links, and provenance (Appendix~\ref{sec:synthesis_details}).

\subsubsection{Assertion-Memory Self-Improvement}
\label{sec:rsi_evolution}

After each batch, the main agent summarizes the failure patterns behind joint rejections and evaluator disagreements and formulates candidate assertions. For disagreements, it uses the evaluators' explanations and arbitration outcomes to clarify the underlying criterion, rather than rejecting similar queries merely because evaluators disagreed. Each candidate is compared with existing assertions in scope and requirements; conflicting candidates are revised or withheld, and compatible ones are added to the memory:
\begin{equation}
\mathcal{A}^{(t+1)}
= \operatorname{Update}_{\mathrm{main}}
\left(\mathcal{A}^{(t)},\mathcal{D}^{(t)},\mathcal{J}^{(t)}\right),
\label{eq:rsi_memory_update}
\end{equation}
where $\mathcal{A}^{(t)}$ is the assertion memory at round $t$, and $\mathcal{D}^{(t)}$ and $\mathcal{J}^{(t)}$ collect evaluator disagreements and joint rejections, respectively, together with their feedback. The operator $\operatorname{Update}_{\mathrm{main}}$ denotes failure summarization, assertion formulation, and compatibility assessment by the main agent.

Subsequent queries are checked against the updated memory, so recurring defects are caught before finalization without passing the full assertion history to the generator. Before release, all retained queries, including those accepted in early rounds, are rechecked against the final memory $\mathcal{A}^{(T)}$; queries that fail are repaired or removed.

\subsection{Dataset Composition and Synthesis Quality}
\label{sec:stats}

\subsubsection{Regional composition.}
Table~\ref{tab:dataset-stats}(a) summarizes each region's knowledge
base and queries.
The regions differ sharply in POI density: Chaoyang contains over 150
times as many active POIs as Dabancheng.
Huangpu has the most constraints per query and the highest
constraint-combination entropy, while Chaoyang leads in intent entropy
and semantic distance.
The regions thus pose complementary challenges: Huangpu stresses
compositional constraint satisfaction, whereas Chaoyang stresses
coverage of diverse intents.

\subsubsection{Constraint coverage.}
The benchmark contains 1,200 queries with 3,049 atomic constraint
annotations across 20 dimensions; 75\% of queries involve multiple
constraints, and 196 (16.3\%) are unsatisfiable.
These unsatisfiable queries contain conflicting hard constraints and
test whether an agent reports that no candidate meets all hard
constraints, rather than returning an unsuitable POI.
Table~\ref{tab:dataset-stats}(b) reports, for each region, the share
of queries containing each constraint type; since a query may combine
several types, the shares do not sum to 100\%.
Appendices~\ref{app:constraint-task-distributions}
and~\ref{sec:metric_definitions} report dimension-level distributions,
diversity-metric definitions, and a comparison with LocalSearchBench.

\begin{table}[!ht]
    \centering
    \caption{Regional statistics and constraint coverage.
    A query may contain multiple constraint types.}
    \label{tab:dataset-stats}
    \label{tab:regional_diversity}
    \label{tab:taxonomy}

    \begingroup
    \small
    \setlength{\tabcolsep}{0pt}
    \renewcommand{\arraystretch}{0.95}
    \setlength{\aboverulesep}{1pt}
    \setlength{\belowrulesep}{1pt}
    \arrayrulecolor{AtlasTrajectory!55}

    \begin{tabular*}{\linewidth}
        {@{\extracolsep{\fill}}lccc@{}}
        \toprule
        \textbf{Metric}
        & \textbf{Chaoyang}
        & \textbf{Huangpu}
        & \textbf{Dabancheng} \\
        \midrule
        \multicolumn{4}{@{}l}{\textit{(a) Corpus statistics}} \\
        Queries
        & 400 & 400 & 400 \\
        Active POIs
        & 343,916 & 79,058 & 2,270 \\
        Mean constraints/query
        & 2.17 & 3.09 & 2.37 \\
        Intent entropy
        & 0.8329 & 0.6264 & 0.6750 \\
        Constraint combination entropy
        & 0.6608 & 0.8130 & 0.5788 \\
        Surface diversity
        & 0.4067 & 0.4327 & 0.2854 \\
        Mean cosine distance
        & 0.5898 & 0.5630 & 0.5338 \\
        Inter-annotator agreement
        & 0.924 & 0.925 & 0.917 \\
        \midrule
        \multicolumn{4}{@{}l}{
            \textit{(b) Query coverage by constraint type (\%)}
        } \\
        Hard objective
        & 64.3 & 82.5 & 71.3 \\
        Hard subjective
        & 12.8 & 13.5 & 26.3 \\
        Soft objective
        & 16.5 & 52.0 & 27.0 \\
        Soft subjective
        & 53.0 & 25.0 & 41.8 \\
        \bottomrule
    \end{tabular*}
    \endgroup
    \vspace{-3pt}
\end{table}

\subsubsection{Synthesis quality across RSI rounds.}
\label{sec:rsi-quality}
\begin{figure}[htbp]
    \centering
    \includegraphics[width=\columnwidth]{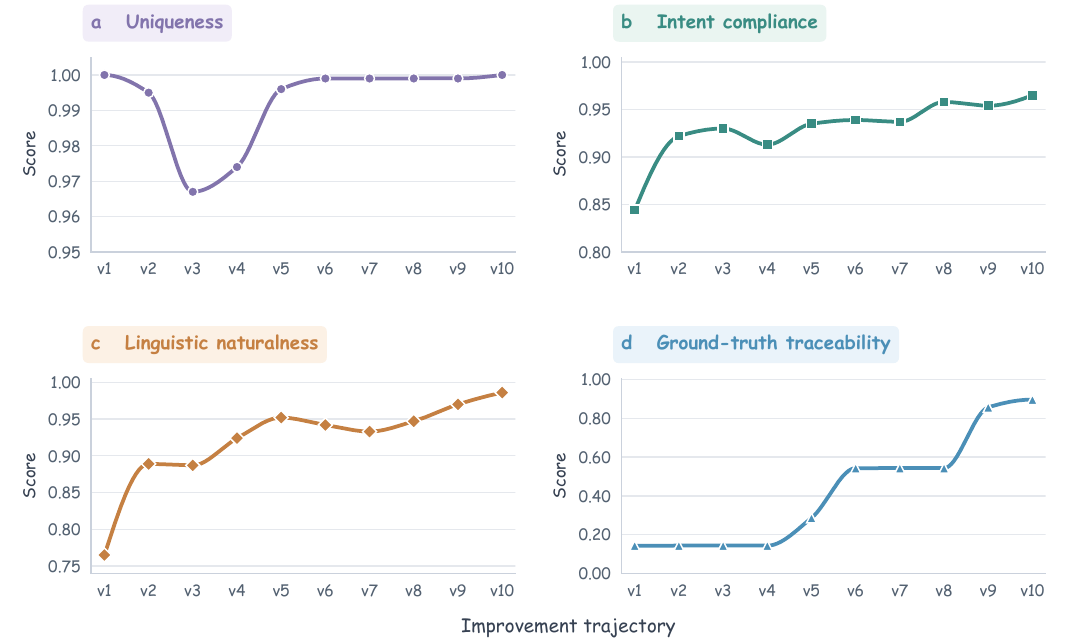}
    % \vspace{-4pt}
    \caption{Observed synthesis-quality trajectories across
    ten RSI rounds.}
    % : (a) uniqueness, (b) intent compliance,
    % (c) linguistic naturalness, and (d) reference-target
    % traceability. }
    \label{fig:rsi-improvement}
    \vspace{-14pt}
\end{figure}

Figure~\ref{fig:rsi-improvement} tracks synthesis quality
across ten recursive improvement rounds.
Disagreements and joint rejections from Debate informed
assertion-memory updates, which the main agent assessed
for compatibility before subsequent use at Assert.
Recorded revisions include corrections to evidence linkage,
refinements to constraint checks, and relaxation of
over-restrictive rejection rules
(Appendix~\ref{app:assertion-evolution}).
From the first to the tenth round, intent compliance
increased from 0.844 to 0.965, linguistic naturalness
from 0.765 to 0.986, and reference-target traceability
from 0.143 to 0.897.
Uniqueness remained at or above 0.967 and reached 1.000
in the final round.
Intermediate declines indicate that improvement was
not monotonic.

% \begin{figure}[htbp]
%     \centering
%     \includegraphics[width=\columnwidth]{pic/rsi_trj.pdf}
%     \caption{Observed synthesis-quality trajectories across
%     ten RSI rounds: (a) uniqueness, (b) intent compliance,
%     (c) linguistic naturalness, and (d) reference-target
%     traceability. Panels use different y-axis ranges.}
%     \label{fig:rsi-improvement}
% \end{figure}

% \subsection{Integration of ATLASBox}
% \label{sec:sandbox}

% ATLASBox turns each benchmark record into an executable
% local-search task. The agent receives the query, timestamp,
% and user coordinates; raw persona fields and reference-target
% annotations are withheld. It can call four standardized tools
% for POI search, review search, POI-specific reviews, and
% radius-constrained retrieval (Appendix~\ref{app:tools}).
% Structured attributes and reviews are
% served from fixed regional knowledge bases to ensure that different
% agents operate over the same underlying information. Each returned
% record is associated with evidence identifiers, allowing subsequent
% claims and recommendations to be traced back to the retrieved
% attributes or review content.

% The sandbox records tool arguments, observations, parallel
% calls, final recommendations, latency, and token usage.
% Answer evaluation checks constraints and evidence without
% treating the clicked POI as a unique correct answer.
% Trajectory evaluation examines request interpretation, tool
% selection, evidence verification, and adaptation to observations.
% This setup supports comparable model runs and diagnoses errors
% in retrieval, reasoning, and final decisions.

\begin{table*}[t]
    \centering
    \caption{Overall performance on ATLAS-Bench.
    The \protect\colorbox{AtlasTrajectory!18}{\strut\textbf{best}}
    and \protect\colorbox{AtlasGreen!15}{\strut\underline{second-best}}
    values are highlighted, excluding efficiency.}
    \label{tab:overall-performance}

    \begingroup
    % Keep model names and reasoning settings on the same line.
    \renewcommand{\AtlasModel}[2]{%
        \mbox{#1~{\scriptsize\color{AtlasMuted}(#2)}}}

    % Highlight the largest and second-largest values,
    % excluding efficiency columns.
    \newcommand{\AtlasMax}[1]{%
        \begingroup\setlength{\fboxsep}{1.2pt}%
        \colorbox{AtlasTrajectory!18}{\strut\textbf{#1}}\endgroup}
    \newcommand{\AtlasSecond}[1]{%
        \begingroup\setlength{\fboxsep}{1.2pt}%
        \colorbox{AtlasGreen!15}{\strut\underline{#1}}\endgroup}

    % Use a lighter text color for all efficiency values.
    \newcommand{\AtlasCost}[1]{%
        \textcolor{AtlasInk!65}{#1}}

    \setlength{\tabcolsep}{3.2pt}
    \renewcommand{\arraystretch}{1.16}
    \setlength{\heavyrulewidth}{0.7pt}
    \setlength{\lightrulewidth}{0.4pt}
    \arrayrulecolor{AtlasRule}
    \color{AtlasInk}
    \footnotesize

    \resizebox{\textwidth}{!}{%
    \begin{tabular}{l*{12}{c}}
        \toprule
        & \multicolumn{5}{>{\columncolor{AtlasAnswerBg}}c}{
            \rule[-4pt]{0pt}{16pt}
            \textcolor{AtlasAnswer}{\bfseries Answer Quality}}
        & \multicolumn{4}{>{\columncolor{AtlasTrajectoryBg}}c}{
            \textcolor{AtlasTrajectory}{\bfseries Trajectory Quality}}
        & \multicolumn{3}{>{\columncolor{AtlasEfficiencyBg}}c}{
            \textcolor{AtlasEfficiency}{\bfseries Efficiency}} \\
        \addlinespace[4pt]
        \makecell[l]{\bfseries Model}
        & \AtlasHead{CSR}
        & \AtlasHead{CR}
        & \AtlasHead{Success@k}
        & \AtlasHead{Recs/q}
        & \AtlasHead{Factual}
        & \AtlasHead{Intent}
        & \AtlasHead{Approp.}
        & \AtlasHead{Evidence}
        & \AtlasHead{Strategy}
        & \AtlasHead{Tokens/q}
        & \AtlasHead{Calls/q}
        & \AtlasHead{Rounds/q} \\
        \midrule

        \AtlasFamily{Qwen3.5-4B}
        \hspace{6pt} w/o Think
        & 37.3 & 65.8 & 54.0 & 2.24 & 40.25
        & 3.53 & 3.70 & 1.99 & 2.96
        & \AtlasCost{36.8K} & \AtlasCost{3.79} & \AtlasCost{1.92} \\
        \hspace{6pt} w/ Think
        & 44.0 & 71.4 & 57.7 & 1.81 & 54.33
        & 3.17 & 3.84 & 2.11 & 3.19
        & \AtlasCost{50.3K} & \AtlasCost{4.74} & \AtlasCost{2.73} \\
        \addlinespace[2pt]

        \AtlasFamily{Qwen3.6-35B-A3B}
        \hspace{6pt} w/o Think
        & 46.5 & 70.9 & 73.1 & 2.43 & 49.83
        & 3.93 & 3.99 & 2.27 & 3.50
        & \AtlasCost{57.6K} & \AtlasCost{6.07} & \AtlasCost{3.33} \\
        \hspace{6pt} w/ Think
        & 54.3 & 77.0 & 73.9 & 2.05 & 67.25
        & 3.62 & 4.03 & 2.37 & 3.36
        & \AtlasCost{56.3K} & \AtlasCost{5.15} & \AtlasCost{2.97} \\
        \addlinespace[2pt]

        \AtlasFamily{Qwen3.8-27B}
        \hspace{6pt} w/o Think
        & 55.0 & 73.9 & 75.4 & 2.74 & 77.83
        & 4.19 & 4.16 & 2.40 & 3.43
        & \AtlasCost{46.8K} & \AtlasCost{4.28} & \AtlasCost{2.27} \\
        \hspace{6pt} w/ Think
        & 61.2 & 77.9 & 82.9 & 2.98 & 78.42
        & 4.10 & \AtlasSecond{4.26} & \AtlasSecond{2.61} & 3.71
        & \AtlasCost{73.3K} & \AtlasCost{5.79} & \AtlasCost{3.11} \\
        \addlinespace[3pt]

        \rowcolor{AtlasStripe}
        \AtlasModel{Gemini-3.8-Flash}{Hidden Think}
        & \AtlasSecond{68.3} & \AtlasMax{85.2} & 76.2 & 1.88 & 79.08
        & 4.04 & 4.19 & 2.50 & 3.37
        & \AtlasCost{45.9K} & \AtlasCost{4.73} & \AtlasCost{2.48} \\

        \AtlasModel{Claude-5-Sonnet}{Adaptive}
        & 62.2 & 78.5 & 80.8 & 1.82 & 77.67
        & 3.64 & 4.22 & 2.57 & 3.50
        & \AtlasCost{65.4K} & \AtlasCost{4.56} & \AtlasCost{2.84} \\

        \rowcolor{AtlasStripe}
        \AtlasModel{Kimi-K3}{Max}
        & 55.9 & 74.3 & 82.4 & 2.83 & 79.33
        & 3.74 & 4.25 & 2.56 & 3.63
        & \AtlasCost{56.0K} & \AtlasCost{5.49} & \AtlasCost{3.04} \\

        \AtlasModel{GPT-5.6-Luna}{Medium}
        & 67.7 & 81.2 & 69.0 & 1.61 & \AtlasSecond{90.42}
        & 4.01 & 4.17 & 2.40 & 3.44
        & \AtlasCost{53.7K} & \AtlasCost{4.20} & \AtlasCost{2.56} \\

        \rowcolor{AtlasStripe}
        \AtlasModel{GPT-6-Luna}{XHigh}
        & \AtlasMax{71.5} & \AtlasSecond{84.3} & 67.3 & 1.25
        & \AtlasMax{95.58}
        & 4.23 & 4.16 & 2.41 & 3.31
        & \AtlasCost{74.8K} & \AtlasCost{5.11} & \AtlasCost{3.21} \\

        \AtlasModel{Qwen3.8-Max}{Max}
        & 58.3 & 76.1 & 84.0 & \AtlasSecond{3.09} & 83.00
        & \AtlasMax{4.30} & \AtlasSecond{4.26} & 2.57 & 3.90
        & \AtlasCost{79.2K} & \AtlasCost{6.47} & \AtlasCost{3.46} \\

        \rowcolor{AtlasStripe}
        \AtlasModel{DeepSeek-V4.1-Flash}{Thinking}
        & 55.9 & 74.2 & \AtlasSecond{84.7} & \AtlasMax{3.38} & 80.83
        & \AtlasSecond{4.28} & 4.20 & \AtlasMax{2.62}
        & \AtlasSecond{3.94}
        & \AtlasCost{94.7K} & \AtlasCost{7.39} & \AtlasCost{3.82} \\

        \AtlasModel{GLM-5.3}{Max}
        & 56.9 & 75.1 & \AtlasMax{85.3} & 3.07 & 83.67
        & 4.25 & \AtlasMax{4.35} & \AtlasSecond{2.61} & \AtlasMax{4.05}
        & \AtlasCost{90.8K} & \AtlasCost{7.04} & \AtlasCost{3.64} \\

        \bottomrule
    \end{tabular}%
    }
    \endgroup
\end{table*}

\subsection{Integration of ATLASBox}
\label{sec:sandbox}

ATLASBox turns each benchmark record into an executable local-search
task in a standardized environment.
The agent receives only the query, timestamp, and user coordinates;
persona fields and reference-target annotations are withheld to prevent
information leakage.
It can call four tools for POI search, review search, POI-specific
review lookup, and radius-constrained retrieval
(Appendix~\ref{app:tools}), all served from fixed regional knowledge
bases so that every agent accesses the same information.
Each returned record carries an evidence identifier, so claims and
recommendations can be traced back to the retrieved attributes or reviews.

The sandbox logs tool arguments, observations, parallel calls, final
recommendations, latency, and token usage.
Answer evaluation checks whether returned POIs satisfy the annotated
constraints and whether supporting claims are grounded in retrieved
evidence, without treating the clicked POI as the only correct answer.
Trajectory evaluation examines request interpretation, tool selection,
evidence verification, and adaptation to intermediate observations.
Together, these enable comparable model runs and fine-grained diagnosis
of failures in retrieval, reasoning, tool use, and final decisions.

\section{Experiment}

\subsection{Metrics}
Using evidence from the recommended POI, we classify each recommendation
as \emph{qualified} if all hard constraints and soft preferences are
supported; \emph{acceptable} if all hard constraints are satisfied and
any unmet or unverifiable soft preferences are explicitly disclosed;
or \emph{unacceptable} if a hard constraint is violated or unsupported,
a soft-preference gap is undisclosed, or evidence is fabricated,
distorted, or drawn from another POI.

% \textit{1. Answer Quality.}
% Based on these verdicts, we evaluate answer quality using five
% complementary metrics.
% \textbf{CSR} is a strict measure of \textbf{c}onstraint \textbf{s}atisfaction
% at the \textbf{r}ecommendation level.
% It credits satisfied constraints only in qualified or
% acceptable recommendations, while constraints in
% unacceptable recommendations receive no credit.
% \textbf{CR} instead measures constraint recall: it counts every requested
% constraint that is successfully covered by supporting evidence, including
% constraints recalled within an otherwise unacceptable recommendation,
% and therefore gives partial credit to such recommendations.
% \textbf{Success@k} measures the proportion of should-answer queries for
% which at least one qualified or acceptable recommendation is returned,
% so empty answers are counted as failures and over-refusal is penalized.
% \textbf{Recs/query} averages the number of recommendations over the same
% queries. Empty answers contribute zero to both metrics.
% Finally, \textbf{factual accuracy} requires passing both a deterministic
% grounding check and an LLM-based factuality check.
\textit{1. Answer Quality.}
Based on these verdicts, we report five complementary metrics, formally defined in Appendix~\ref{app:answer-quality-metrics}.
\textbf{CSR} strictly measures \textbf{c}onstraint \textbf{s}atisfaction
at the \textbf{r}ecommendation level, crediting satisfied constraints
only in qualified or acceptable recommendations.
\textbf{CR} measures \textbf{c}onstraint \textbf{r}ecall: it credits every
evidence-supported constraint, including those in unacceptable
recommendations, which thus receive partial credit.
\textbf{Success@k} is the proportion of should-answer queries with at least
one qualified or acceptable recommendation, and \textbf{Recs/query} is the
average number of recommendations over the same queries. Empty answers
count as zero in both, so over-refusal is penalized.
\textbf{Factual accuracy} requires passing both a deterministic grounding
check and an LLM-based factuality check.
\textit{2. Trajectory Quality.}
We assess \textbf{intent understanding},
\textbf{tool-use appropriateness}, \textbf{evidence sufficiency} and  \textbf{strategy adaptation}.
\textit{3. Execution Cost.}
We report \textbf{token consumption}, \textbf{tool calls}, and \textbf{execution rounds}
per query to characterize the resources required by each model.
\subsection{Experimental Setup}
\label{sec:experimental-setup}

\textit{Models.}
% We evaluate Qwen3.8-Max~\cite{qwen38}, Qwen3.5-4B~\cite{qwen35blog},Qwen3.8-27B, Claude-5-Sonnet~\cite{anthropic_claude_sonnet_5}, GPT-5.6-Luna~\cite{openai_gpt56},
% GLM-5.3~\cite{glm5team2026glm5vibecodingagentic}, Kimi-K3~\cite{kimiteam2026kimik3openfrontier}, DeepSeek-V4.1-flash~\cite{deepseekai2026deepseekv41flashpushinglimitskv}, and Gemini-3.8-Flash~\cite{google_gemini_38_flash} in ATLASBox,
% using the same knowledge bases and tool interfaces. Appendix~\ref{app:tools} documents the tools.
We evaluate 14 configurations of 11 models:
Qwen3.5-4B~\cite{qwen35blog},
Qwen3.6-35B-A3B~\cite{qwen36_35b_a3b},
Qwen3.8-27B and Qwen3.8-Max~\cite{qwen38},
Claude-5-Sonnet~\cite{anthropic_claude_sonnet_5},
GPT-5.6-Luna~\cite{openai_gpt56},
GPT-6-Luna~\cite{openai_gpt6_luna},
GLM-5.3~\cite{glm5team2026glm5vibecodingagentic},
Kimi-K3~\cite{kimiteam2026kimik3openfrontier},
DeepSeek-V4.1-Flash~\cite{deepseekai2026deepseekv41flashpushinglimitskv},
and Gemini-3.8-Flash~\cite{google_gemini_38_flash}.
The first three Qwen models are tested with and without
thinking. 
\textit{Settings.}
All models share ATLASBox's knowledge bases
and tools (Appendix~\ref{app:tools}). Retrieval uses Qwen3-Embedding-0.6B
to retrieve 100 candidates, followed by
Qwen3-Reranker-0.6B with adaptive top-$k$
(default 10, maximum 20).
Agents are limited to $N=5$ tool-use rounds.
Local computation uses two NVIDIA H20 GPUs.
% \paragraph{Answer Quality.}
% We evaluate answer quality along five complementary dimensions.
% CSR measures strict constraint satisfaction by crediting satisfied constraints only in
% qualified or acceptable recommendations, while CR additionally gives partial credit to
% constraints satisfied by otherwise unacceptable recommendations.
% Success@k measures whether at least one usable recommendation is returned for a
% satisfiable query, thereby penalizing over-refusal, and Recs/query captures the average
% number of recommendations surfaced on such queries.
% Diversity measures the semantic spread of returned POIs using conditional ILD@10.
% Finally, factual accuracy requires answers to be both correctly grounded in retrieved
% evidence and free of factual errors.
% Together, these metrics capture constraint fulfillment, answer coverage, recommendation
% breadth, diversity, and evidence-grounded factuality.
\textit{Judge.}
All LLM-based judgments use DeepSeek-V4-Pro~\cite{deepseekai2026deepseekv4}.
Since computing CSR and CR requires checking each recommended POI
against the query constraints and its supporting evidence, we adopt
the Agent-as-a-Judge paradigm~\cite{zhuge2024agent}: the judge uses
tools to locate, retrieve, and inspect evidence and verifies each
constraint individually, reducing grounding and verification errors.
We implement this workflow in Qoder~\cite{qoder2025}, a coding-agent environment that,
like Claude Code and Codex, supports tool use and iterative evidence
inspection. With DeepSeek-V4-Pro as the backbone, we package the
evaluation protocol as reusable agent skills~\cite{li2026skillsbench}
and parallelize checks across subagents.
Other LLM-based evaluations call DeepSeek-V4-Pro directly via API. 

\subsection{Overall Performance and Analysis}
\label{sec:overall-performance}
\subsubsection{Answer Quality Analysis.}
As shown in Table~\ref{tab:overall-performance}, GPT-6-Luna achieves
the highest CSR (71.5\%) and factual accuracy (95.58\%) with a high CR
(84.3\%), indicating precise, well-grounded recommendations.
However, its reluctance to relax soft preferences yields only 1.25
recommendations per query and a Success@k of just 67.3\%.
In contrast, GLM-5.3 and DeepSeek-V4.1-Flash return about three
recommendations per query and reach higher Success@k (85.3\% and
84.7\%), but trail in CSR, CR, and factual accuracy.
This reveals a  trade-off: conservative models offer higher
precision and factual reliability but lower coverage, whereas more
exploratory models improve answer coverage and recommendation
breadth at the cost of including more imperfect candidates.
The five metrics should therefore be interpreted jointly.
Among open-source models, Qwen3.8-27B with thinking best balances
constraint satisfaction and answer coverage, and thinking improves all
five answer-quality metrics over its non-thinking variant. Although compact or sparsely activated models are important
for local-service applications with stringent latency requirements,
the weaker performance of Qwen3.5-4B and Qwen3.6-35B-A3B suggests
that targeted post-training may be needed to improve their
constraint satisfaction and evidence grounding. 
% we also validate the judge labels (\textit{qualified}, \textit{acceptable}, or \textit{unacceptable}) underlying CSR, CR, and Success@k against human expert annotations on 201 recommended POIs, achieving 93.03\% exact agreement (Cohen's $\kappa = 0.891$).
% Across 201 sampled recommended POIs, the \textit{qualified}/\textit{acceptable}/\textit{unacceptable} labels underlying CSR, CR, and Success@k match human expert annotations with 93.03\% exact agreement (Cohen's $\kappa = 0.891$).
Across 201 sampled recommended POIs, the judge labels (\textit{qualified}, \textit{acceptable}, or \textit{unacceptable}) underlying CSR, CR, and Success@k match human expert annotations with 93.03\% exact agreement (Cohen's $\kappa = 0.891$).

% As a consistency check, although we do not restrict valid answers to the single logged reference POI, reference-POI hit rate yields nearly identical model rankings to Success@k (Spearman $\rho=0.93$).

% \paragraph{Recommendation breadth and answer reliability.}
% GPT-6-Luna achieves the highest CSR (71.5\%) and factual
% accuracy (95.58\%), but lower Success@k than GLM-5.3
% (67.3\% versus 85.3\%). They return 1.25 and 3.07
% recommendations per query, respectively, consistent with
% different tendencies toward selective recommendations and
% broader candidate coverage. More candidates may increase
% the chance of finding a valid result while adding constraint
% verification and evidence-grounding demands. Success@k
% should therefore be interpreted alongside recommendation
% count and factual accuracy.

% \paragraph{Gains and costs of thinking.}
% Enabling thinking improves CSR by approximately 6--8
% percentage points across the three paired Qwen models,
% potentially reflecting more thorough constraint verification.
% However, Qwen3.6-35B-A3B gains only 0.8 percentage points
% in Success@k while its median latency increases 8.9-fold,
% despite fewer tool calls. The additional latency cannot
% be explained by retrieval frequency alone and may instead
% reflect greater internal reasoning costs. Longer reasoning
% does not necessarily yield more feasible candidates.

\vspace{-3pt}
\subsubsection{Trajectory Quality Analysis.}
Strong answer quality does not imply uniformly strong trajectory
quality, as different models exhibit distinct strengths.
Qwen3.8-Max leads in intent understanding (4.30), GLM-5.3 in tool-use
appropriateness (4.35) and strategy adaptation (4.05), and
DeepSeek-V4.1-Flash in evidence sufficiency (2.62), while Qwen3.8-27B
with thinking remains competitive across all four dimensions among
open-source models.
GPT-6-Luna pairs strong intent understanding (4.23) with the highest
factual accuracy, but scores lower in evidence sufficiency (2.41) and
strategy adaptation (3.31), suggesting that it gathers limited evidence
and rarely explores alternative strategies.
This may explain its high refusal rate and low Success@k: its answers
are accurate, but more queries go unresolved.
Trajectory and answer metrics thus offer complementary views of model
behavior.
To validate these automated trajectory scores, we compared them with
human expert ratings on a stratified sample of 100 trajectories:
linearly weighted Cohen's $\kappa$ was 0.837 for intent understanding,
0.729 for tool-use appropriateness, 0.891 for evidence sufficiency,
and 0.832 for strategy adaptation, with exact agreement ranging from
81\% to 88\%. 
\begin{figure}[t]
    \centering
    \includegraphics[width=\columnwidth]
        {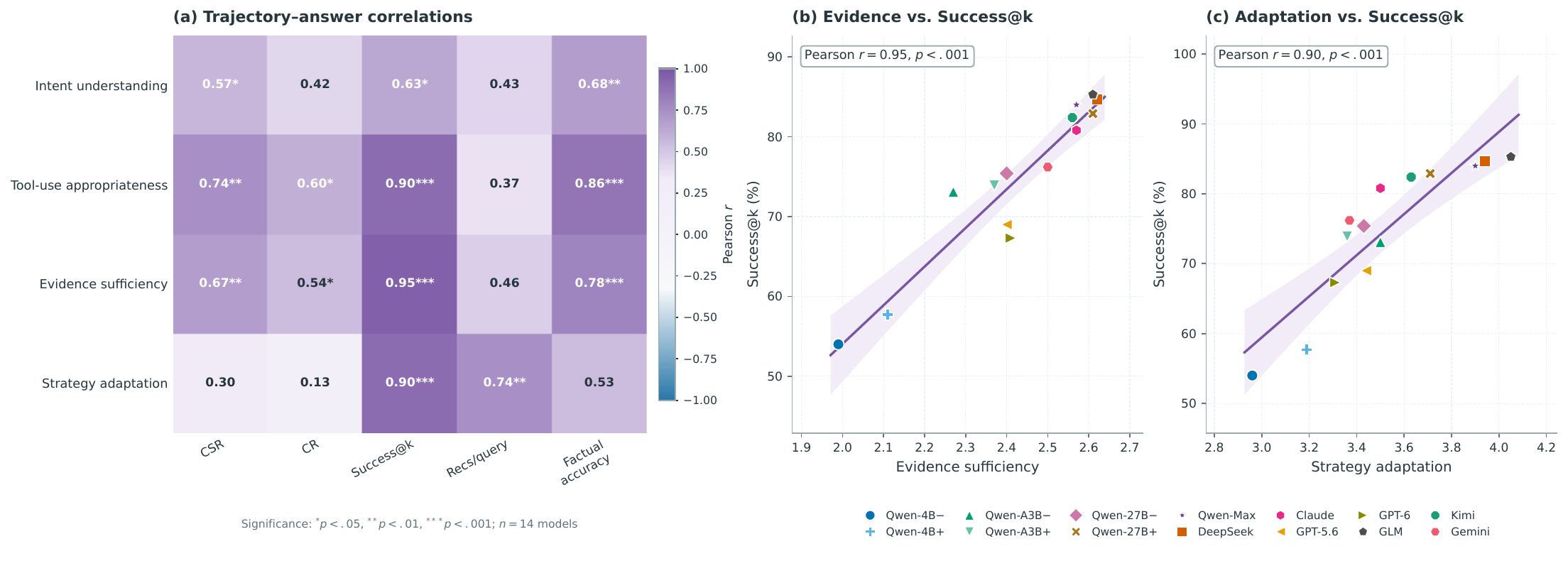}
    \caption{Correlations between trajectory and answer quality.}
    \label{fig:trajectory-answer-correlation}
\end{figure}

% As shown in Figure~\ref{fig:trajectory-answer-correlation}, trajectory quality is closely associated with query success across the 14 evaluated model configurations. Tool-use appropriateness, evidence sufficiency, and strategy adaptation correlate strongly with Success@k ($r=0.90$, $0.95$, and $0.90$, respectively). Strategy adaptation also correlates with recommendation volume (Recs/query; $r=0.74$), whereas intent understanding is more closely associated with factual accuracy ($r=0.68$).

As shown in Figure~\ref{fig:trajectory-answer-correlation}, tool-use
appropriateness, evidence sufficiency, and strategy adaptation
correlate strongly with Success@k across the 14 configurations
($r=0.90$, $0.95$, and $0.90$). Strategy adaptation also correlates
with Recs/query ($r=0.74$), while intent understanding aligns with
factual accuracy ($r=0.68$). However, excluding the Qwen series (where
scaling lifts all metrics), evidence sufficiency and strategy
adaptation correlate negatively with CSR across the remaining seven
models ($r=-0.89$ and $-0.86$), showing that broader exploration finds
more answers but admits more constraint violations.

\subsubsection{Efficiency.}
Qwen3.5-4B without thinking uses the fewest tokens (36.8K),
tool calls (3.79), and rounds (1.92) per query.
DeepSeek-V4.1-Flash consumes the most resources, averaging
94.7K tokens, 7.39 tool calls, and 3.82 rounds.
Its high Success@k comes with greater execution cost,
highlighting a trade-off between query success and efficiency.

\subsection{Regional Analysis}
\label{sec:poi-density}

\begin{figure}[htbp]
    \centering
    \includegraphics[width=\columnwidth]{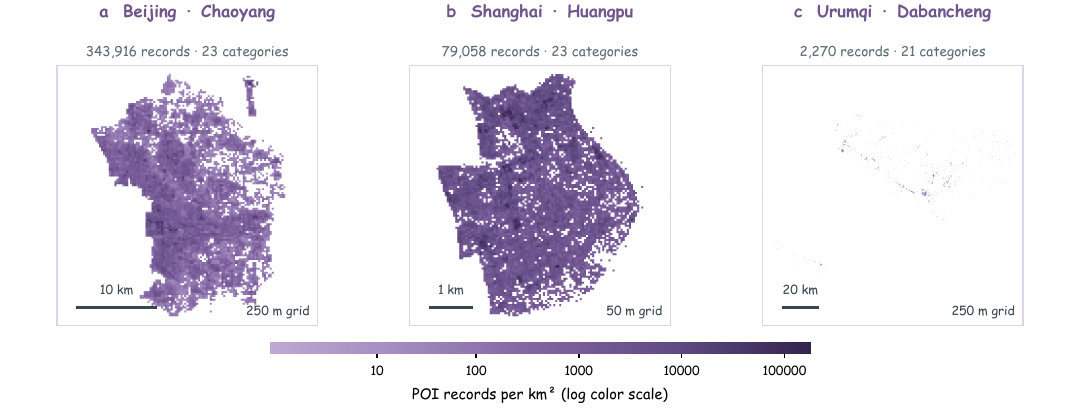}
    \caption{POI distributions, colored by
area-normalized density on a shared log scale.
}
    \label{fig:poi-distrubution}
    % \vspace{-8pt}
\end{figure}

Figure~\ref{fig:poi-distrubution} illustrates the contrasting
POI distributions in Chaoyang, Huangpu, and Dabancheng,
providing settings for examining how spatial density
relates to agentic search performance. Model rankings are consistent across regions, with Spearman
correlations of $0.88$ to $0.90$ for Success@k and $0.82$ to
$0.91$ for evidence sufficiency.
Despite this ranking stability, regional averages reveal
substantial differences in performance and cost
(Table~\ref{tab:regional-performance}).
Huangpu has the lowest CSR (41.4\%) and Success@k (63.7\%),
while requiring the most tool calls (6.12) and tokens (75.4K)
per query.
Its evidence sufficiency also remains lowest (1.90),
suggesting that greater search effort does not necessarily
produce adequate supporting evidence.
The combination of dense candidates and more complex queries
may make candidate selection and evidence verification
particularly demanding.
Dabancheng achieves a higher Success@k (80.8\%) with only
3.88 tool calls and 43.0K tokens per query, consistent with
a smaller candidate pool reducing search effort.
However, its correct-refusal rate on unsatisfiable queries
is lower than Chaoyang's (75.6\% versus 87.5\%).
Sparse candidate availability may therefore pose a different
challenge: agents may retrieve fewer POIs but still select
an unsuitable candidate to provide an answer, rather than
recognize that none satisfies the constraints.
This suggests that efficient retrieval and reliable recognition
of unsatisfiable queries remain distinct capabilities. 
\begin{table}[htbp]
    \centering
    \caption{Regional performance across 14 configurations.}
    \label{tab:regional-performance}

    \begingroup
    \footnotesize
    \setlength{\tabcolsep}{3.5pt}
    \renewcommand{\arraystretch}{1.22}
    \arrayrulecolor{AtlasRule}
    \color{AtlasInk}

    \newcommand{\RegionalHighlight}[1]{%
        \cellcolor{AtlasTrajectory!18}\textbf{#1}}

    \resizebox{\columnwidth}{!}{%
    \begin{tabular}{@{}l*{8}{c}@{}}
        \toprule
        & \multicolumn{5}{>{\columncolor{AtlasAnswerBg}}c}
            {\textcolor{AtlasAnswer}{\bfseries Answer Quality}}
        & \multicolumn{1}{>{\columncolor{AtlasTrajectoryBg}}c}
            {\textcolor{AtlasTrajectory}{\bfseries Trajectory}}
        & \multicolumn{2}{>{\columncolor{AtlasEfficiencyBg}}c}
            {\textcolor{AtlasEfficiency}{\bfseries Efficiency}} \\
        \addlinespace[3pt]
        \textbf{Region}
        & CSR
        & CR
        & Success@k
        & \makecell{Recs\\(/query)}
        & \makecell{Factual\\accuracy}
        & \makecell{Evidence\\sufficiency}
        & \makecell{Tool\\calls}
        & Tokens \\
        \midrule
        Chaoyang
& 62.9 & 77.1 & 80.0 & \RegionalHighlight{2.81}
& 72.6 & 2.46 & 6.04 & 70.5K \\
        \addlinespace[2pt]
        \rowcolor{AtlasStripe}
        Huangpu
        & 41.4 & 68.8 & 63.7 & 2.08
        & 68.6 & 1.90 & \RegionalHighlight{6.12} & \RegionalHighlight{75.4K} \\
        \addlinespace[2pt]
        Dabancheng
        & \RegionalHighlight{68.7}
        & \RegionalHighlight{84.0}
        & \RegionalHighlight{80.8}
        & 2.22
        & \RegionalHighlight{81.2}
        & \RegionalHighlight{2.91}
        & {3.88}
        & {43.0K} \\
        \bottomrule
    \end{tabular}%
    }
    \endgroup
\end{table}
\subsection{Ablation Analysis}

\begin{table*}[t]
    \centering
    \caption{Budget sensitivity of Qwen3.8-27B (w/o thinking).
    $C$: candidate-pool size; $T$: results per call;
    $N$: maximum rounds.}
    \label{tab:budget-sensitivity}

    \begingroup
    \scriptsize
    \setlength{\tabcolsep}{3pt}
    \renewcommand{\arraystretch}{1.05}
    \setlength{\aboverulesep}{1pt}
    \setlength{\belowrulesep}{1pt}
    \arrayrulecolor{AtlasRule}
    \color{AtlasInk}

    \begin{tabular*}{\textwidth}{
        @{\extracolsep{\fill}}
        l*{6}{c}|
        *{3}{c}
        >{\columncolor{AtlasGreen!10}}c
        *{3}{c}
        @{}
    }
        \toprule
        & \multicolumn{3}{c}{$C=100$}
        & \multicolumn{3}{c|}{$C=200$}
        & \multicolumn{7}{c}{$N$} \\
        \cmidrule(lr){2-4}
        \cmidrule(lr){5-7}
        \cmidrule(lr){8-14}

        \textbf{Metric}
        & $T=10$ & $T=20$ & $T=30$
        & $T=10$ & $T=20$ & $T=30$
        & 2 & 3 & 4 & \textbf{5} & 6 & 7 & 8 \\
        \midrule

        \textcolor{AtlasAnswer}{CSR (\%)}
        & 56.4 & 60.4 & 56.4 & 59.0 & 59.3 & 53.3
        & 55.9 & 52.8 & 55.8 & 55.0 & 58.3 & 59.1 & 56.8 \\

        \textcolor{AtlasAnswer}{CR (\%)}
        & 75.5 & 74.4 & 74.0 & 76.7 & 77.5 & 73.6
        & 77.2 & 73.5 & 75.5 & 74.0 & 74.1 & 76.1 & 73.6 \\

        \textcolor{AtlasAnswer}{Success@k (\%)}
        & 74.4 & 77.9 & 82.6 & 76.7 & 79.1 & 76.7
        & 73.3 & 75.6 & 79.1 & 80.2 & 80.2 & 81.4 & 77.9 \\

        \textcolor{AtlasAnswer}{Recs/q}
        & 2.24 & 2.47 & 2.73 & 2.34 & 2.48 & 2.71
        & 2.45 & 2.44 & 2.74 & 2.52 & 2.56 & 2.72 & 2.60 \\

        \textcolor{AtlasAnswer}{Factual (\%)}
        & 81.00 & 76.00 & 72.00 & 80.00 & 72.00 & 77.00
        & 79.00 & 73.00 & 79.00 & 77.00 & 76.00 & 78.00 & 76.00 \\
        \midrule

        \textcolor{AtlasTrajectory}{Intent}
        & 4.34 & 4.28 & 4.54 & 4.43 & 4.45 & 4.49
        & 4.30 & 4.42 & 4.27 & 4.29 & 4.51 & 4.50 & 4.39 \\

        \textcolor{AtlasTrajectory}{Approp.}
        & 4.20 & 4.34 & 4.36 & 4.34 & 4.33 & 4.40
        & 4.29 & 4.40 & 4.29 & 4.31 & 4.47 & 4.43 & 4.42 \\

        \textcolor{AtlasTrajectory}{Evidence}
        & 2.59 & 2.68 & 2.91 & 2.75 & 2.76 & 2.59
        & 2.66 & 2.62 & 2.61 & 2.64 & 2.76 & 2.81 & 2.60 \\

        \textcolor{AtlasTrajectory}{Strategy}
        & 3.34 & 3.30 & 3.58 & 3.42 & 3.21 & 3.57
        & 3.17 & 3.41 & 3.44 & 3.45 & 3.53 & 3.52 & 3.57 \\
        \midrule

        \textcolor{AtlasEfficiency}{Tokens/q}
        & 35.9K & 49.4K & 61.0K & 34.3K & 47.9K & 60.0K
        & 29.8K & 34.0K & 40.2K & 42.2K & 49.4K & 49.7K & 53.6K \\

        \textcolor{AtlasEfficiency}{Calls/q}
        & 3.87 & 3.86 & 3.99 & 3.78 & 3.72 & 3.96
        & 2.95 & 3.25 & 3.78 & 3.92 & 4.41 & 4.38 & 4.60 \\

        \textcolor{AtlasEfficiency}{Rounds/q}
        & 2.07 & 2.03 & 2.13 & 2.02 & 2.00 & 2.10
        & 1.55 & 1.71 & 1.99 & 2.08 & 2.31 & 2.31 & 2.42 \\
        \bottomrule
    \end{tabular*}
    \endgroup
\end{table*}

\begin{table}[t]
    \centering
    \caption{Retrieval comparison across candidate-pool sizes $k$.
    Shading marks the default; bold indicates the best value.}
    \label{tab:retrieval-comparison}

    \begingroup
    \scriptsize
    \setlength{\tabcolsep}{2pt}
    \renewcommand{\arraystretch}{1.0}
    \setlength{\aboverulesep}{1pt}
    \setlength{\belowrulesep}{1pt}
    \arrayrulecolor{AtlasRule}
    \color{AtlasInk}

    \begin{tabular*}{\columnwidth}
        {@{\extracolsep{\fill}}lcccccc@{}}
        \toprule
        \rowcolor{AtlasTrajectoryBg}
        \textbf{Method}
        & $\boldsymbol{k}$
        & \textbf{NDCG@10}
        & \textbf{Hit@10}
        & \textbf{Recall@20}
        & \textbf{MRR@20}
        & \textbf{P50 (s)} \\
        \midrule
        BM25 & --
        & 0.606 & 0.813 & 0.632 & 0.689 & 0.059 \\
        Vector & 100
        & 0.664 & 0.881 & 0.675 & 0.771
        & \textbf{0.057} \\
        \midrule
        BM25 + rerank & 100
        & 0.726 & 0.884 & 0.713 & 0.820 & 0.564 \\
        Vector + rerank & 20
        & 0.716 & 0.899 & 0.675 & 0.817 & 0.160 \\
        \rowcolor{AtlasGreen!10}
        Vector + rerank & 100
        & 0.745 & 0.927 & 0.745
        & \textbf{0.833} & 0.568 \\
        Vector + rerank & 200
        & \textbf{0.747} & \textbf{0.935}
        & \textbf{0.757} & 0.831 & 1.078 \\
        \bottomrule
    \end{tabular*}
    \endgroup
\end{table}

Table~\ref{tab:budget-sensitivity} shows that, at $C=100$, increasing the fixed return count from 20
to 30 raises Success@k from 77.9\% to 82.6\%, but reduces
factual accuracy from 76.0\% to 72.0\% and increases
tokens per query from 49.4K to 61.0K.
More results may help uncover feasible candidates but
also introduce distracting evidence, making factual
verification harder.
We therefore retain an adaptive return count capped
at 20 results per call to balance recommendation
success, factual reliability, and cost.

For the round budget, increasing $N$ from 2 to 5 raises
Success@k from 73.3\% to 80.2\%.
Further increases to $N=6,7,8$ yield Success@k of
80.2\%, 81.4\%, and 77.9\%, respectively, while consuming
17--27\% more tokens than $N=5$.
We therefore retain $N=5$ as a practical balance
between recommendation success and execution cost.

Table~\ref{tab:retrieval-comparison} shows that reranking improves
retrieval, but enlarging the candidate pool from $k=100$ to $200$
barely raises NDCG@10 (0.745 to 0.747) while nearly doubling median
latency (0.568 to 1.078\,s).

% Single-column latency figure + analysis paragraph for the Experiments section.
% Usage: \input{latency_analysis.tex}  (requires graphicx; figure floats in one column)
\subsection{Further Analysis}
% \begin{figure}[htbp]
%   \centering
%   \includegraphics[width=\columnwidth]{pic/latency_boxplot_closed (5).pdf}
%   \caption{End-to-end per-query latency of closed-source models (log axis).
%   Bar and diamond mark the median and mean; dots are outliers; numbers above
%   boxes are medians in seconds.}
%   \label{fig:latency-closed}
% \end{figure}
\subsubsection{Gains and Costs of Thinking.}
Enabling thinking improves CSR by approximately 6--8
percentage points across the three paired Qwen models,
potentially reflecting more thorough constraint verification, but it typically increases token consumption and 
latency.
% \subsubsection{Gains and costs of thinking.}
% Enabling thinking raises CSR by 6--8 percentage points for all three
% Qwen models evaluated in both modes, possibly due to more thorough
% constraint verification, but it also increases token consumption and
% latency.

% \paragraph{Latency analysis.}
% Figure~\ref{fig:latency-closed} shows end-to-end per-query latency (log axis;
% 1,200 queries per model). The models split into a fast tier (Gemini-3.8-Flash
% median 25.1\,s, GPT-6-Luna 29.1\,s, GPT-5.6-Luna 33.9\,s, Claude-5-Sonnet
% 47.0\,s) and a slow tier (DeepSeek-V4.1-Flash 81.0\,s, Kimi-K3 115.2\,s,
% GLM-5.3 156.0\,s, Qwen3.8-Max 161.0\,s) that is 3--6$\times$ slower at the
% median with heavier tails (GLM-5.3 P95 497.1\,s). Latency is not simply the
% price of reasoning: GPT-6-Luna, at the highest reasoning effort, is the
% second-fastest model with a tight spread (max 110.4\,s) and the best
% answer-pass rate, whereas GLM-5.3's mean (201.5\,s) far exceeds its median
% (156.0\,s), revealing occasional very long episodes. The sub-50\,s tier meets
% interactive response budgets, while the $>$100\,s tier motivates latency-aware
% model selection or bounded reasoning in deployment.

\subsubsection{Recommendation Richness.}

\begin{figure}[htbp]
    \centering
    \includegraphics[width=\columnwidth]{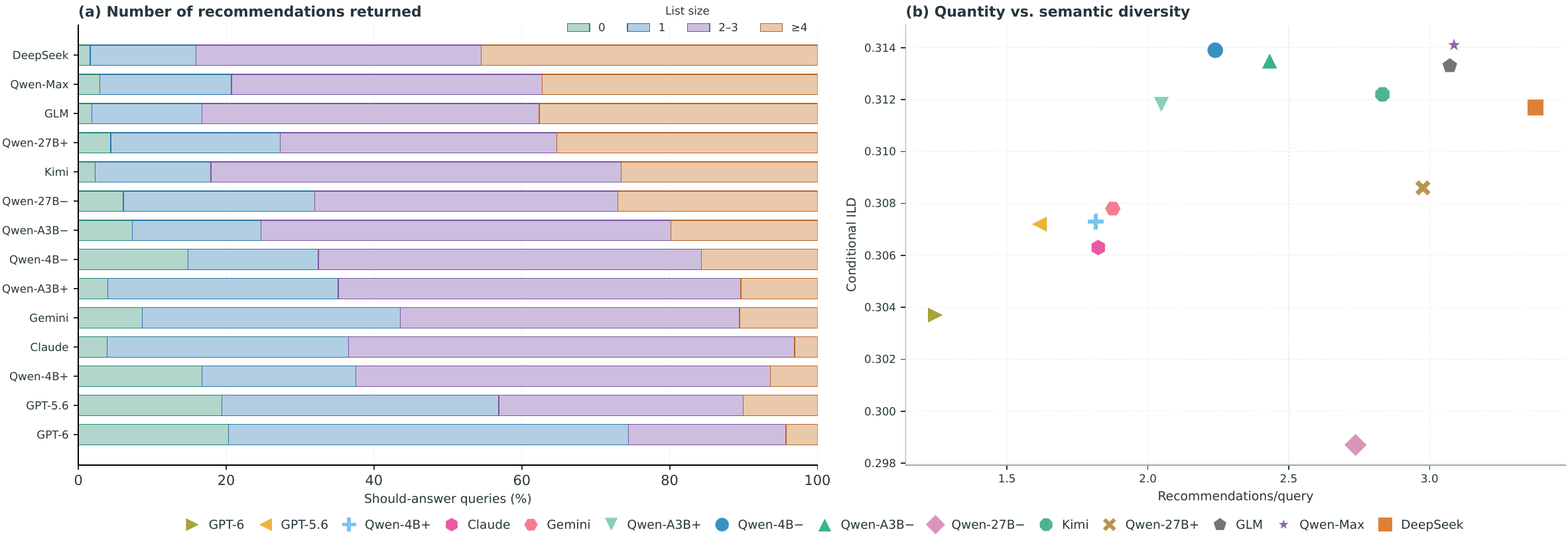}
    \caption{Recommendation richness:
    (a) list-size distributions and
    (b) quantity versus conditional intra-list diversity.}
    \label{fig:recommendation-richness}
    \vspace{-10pt}
\end{figure}

    % Figure~\ref{fig:recommendation-richness} characterizes recommendation
    % richness from two complementary perspectives. Figure~\ref{fig:recommendation-richness}(a) shows the
    % distribution of returned list sizes. DeepSeek, GLM-5.3, and Qwen-Max
    % more frequently provide multiple alternatives, whereas GPT-6-Luna
    % often returns zero or one recommendation, reflecting a more
    % conservative policy. Figure~\ref{fig:recommendation-richness}(b) compares recommendation quantity with
    % conditional intra-list diversity. Qwen-Max and GLM-5.3 combine
    % relatively high recommendation counts with high ILD, while DeepSeek
    % returns the most recommendations (3.38 per query) with slightly lower
    % diversity (0.3117). Recommendation quantity does not necessarily imply
    % diversity: for example, Qwen3.5-4B without thinking returns fewer
    % recommendations but achieves high ILD (0.3139), whereas Qwen3.8-27B
    % without thinking returns more recommendations but has the lowest ILD
    % (0.2987).
    Figure~\ref{fig:recommendation-richness} characterizes this aspect from two perspectives. As shown in
Figure~\ref{fig:recommendation-richness}(a), DeepSeek-V4.1-Flash,
GLM-5.3, and Qwen3.8-Max more often return multiple alternatives,
whereas GPT-6-Luna often returns zero or one, reflecting its
conservative policy.
Figure~\ref{fig:recommendation-richness}(b) relates recommendation
count to intra-list diversity (ILD): Qwen3.8-Max and GLM-5.3 achieve
both, while DeepSeek-V4.1-Flash returns the most recommendations
(3.38 per query) with slightly lower ILD.
Quantity does not imply diversity: models returning more POIs do not
necessarily achieve higher ILD, as their lists may be largely
homogeneous.

\subsubsection{Case Study and Failure Attribution.}
\label{sec:case-study}
\begin{figure}[!t]
    \centering
    \includegraphics[width=\columnwidth]{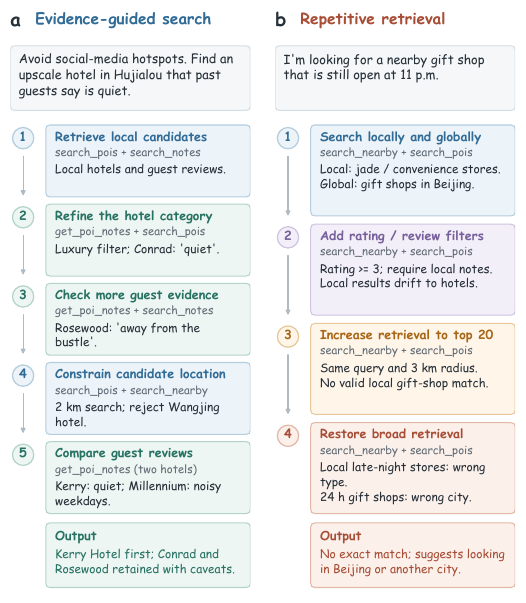}
    \caption{Contrasting tool-use trajectories: (a) evidence-guided
    candidate refinement in a high-scoring case; (b) repetitive,
    off-target retrieval in a failure case.}
    \label{fig:case_studies}
\end{figure}

Figure~\ref{fig:case_studies} contrasts two multi-round tool-use
trajectories to examine how agents use retrieved evidence to revise
their search. In the high-scoring case, filtering by category,
distance, and reviews progressively narrows the candidate set.
Each refinement addresses a requirement of the query, linking
retrieval decisions to the evidence needed for recommendation.
The trajectory illustrates how successive tool calls can contribute
to a more focused search when their results inform the next action.

In the failure case, the agent adjusts filters without resolving
the mismatch between the requested category and locality.
Retrieval consequently remains repetitive and off target, and the
final recommendations violate constraints. The contrast concerns
how the search changes in response to evidence: modifying a filter
does not constitute effective adaptation if the underlying
mismatch persists. These two cases illustrate this distinction;
they do not establish its prevalence across all evaluated models.

Across $1{,}826$ agent-side failures, the reported attribution
assigns $33.7\%$ to intent understanding, $26.2\%$ to tool use,
$22.6\%$ to evidence grounding, and $17.5\%$ to strategy adaptation.
Tool-use errors account for $35.3\%$ of failures among weaker models,
making them the largest category in that group. Intent understanding
is the largest category overall, indicating that failure analysis
must consider the interpretation of requirements alongside tool
execution and evidence use. These percentages describe the
attributed failures, rather than failure rates over all queries.

The cases also help interpret the round-budget results in
Table~\ref{tab:budget-sensitivity}: additional rounds provide
opportunities to inspect evidence, but do not guarantee that an
agent will correct its search. Effective use of the budget requires
identifying which requirement remains unresolved and choosing a
subsequent action that addresses it. This qualitative interpretation
is consistent with the non-monotonic Success@k observed as the
round limit increases beyond five.
% \subsubsection{Evaluation Alignment}
% % On satisfiable queries with reference targets, while valid answers are not limited to the logged POI, reference-target hit rate correlates strongly with Success@k (Spearman $\rho=0.93$).reference-target hit rate correlates strongly with Success@k (Spearman $\rho=0.93$).
% Although valid answers are not restricted to the logged POI, reference-target
% hit rate on satisfiable queries correlates strongly with Success@k at Spearman
% $\rho=0.93$, confirming that our open-ended constraint evaluation aligns
% closely with real user selections.

\subsubsection{Constraint Breakdown.} Across all configurations, Success@k on satisfiable queries with hard objective, hard subjective, soft objective, and soft subjective constraints averages 72.8\%, 60.4\%, 75.9\%, and 69.7\%, likely because subjective constraints are harder to verify than facts while soft preferences allow greater relaxation. 
\subsection{Local Search in Long-Horizon Tasks}
\label{sec:long-horizon-case}
\begin{figure}[htbp]
    \centering
    \includegraphics[width=.90\columnwidth]{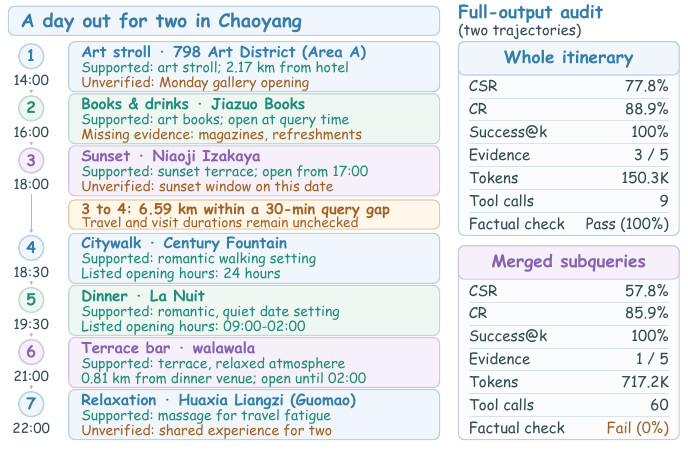}
    % \caption{A long-horizon case showing each stage's top recommendation, evidence, and unresolved checks.}
\caption{Long-horizon case with stage choices and evidence.}

    \label{fig:long-horizon-case}
    % \vspace{-8pt}
\end{figure}

We examine a couple's day-trip request in Beijing's Chaoyang District
with Qwen3.8-Max, comparing a single whole-itinerary execution against
seven merged stage-specific requests.
Figure~\ref{fig:long-horizon-case} shows the first recommendation at
each stage and the evaluation scores of both trajectories. 
While single-pass execution produces a brief, high-level overview with relatively few errors and a CSR of 77.8\%, it offers limited actionable detail and practical guidance for users. 
% decomposing the request into stage-wise subqueries yields a much richer, more actionable itinerary, yet causes CSR to drop to 57.8\% and factuality to fail due to subquery hallucinations.
Decomposing the request into stage-wise subqueries yields
a much richer, more actionable itinerary with greater
practical value for users, but CSR drops to 57.8\%
and the factuality check fails due to evidence-grounding
errors in the subquery outputs.
% The stage-level audit reveals unresolved requirements,
% including magazine availability and a shared relaxation
% experience for two.

This case illustrates why local constraint verification
deserves dedicated evaluation in long-horizon planning:
despite more errors, the subquery-based responses
provide more concrete guidance and greater practical value
for users than direct execution.
This further highlights the importance of our research on reliably solving individual local-search subqueries, ensuring that richer recommendations
are evidence-grounded and satisfy users' constraints. 

\section{Conclusion}
We introduce ATLAS-Bench and ATLASBox for evaluating agentic
local search under diverse user constraints. Our synthesis
pipeline combines privacy rewriting, persona-aware generation,
and iterative validation to ground queries in POI evidence.
Experiments across 11 models with 14 configurations reveal
trade-offs between recommendation coverage and reliability:
higher query success can come with weaker constraint
satisfaction and evidence grounding.
Providing useful options across queries while reliably
meeting users' requirements remains a key challenge.

% \clearpage

\bibliographystyle{ACM-Reference-Format}
\bibliography{main}

\appendix

\section{Data Synthesis Implementation Details}
\label{sec:synthesis_details}

Prepare uses shards of approximately 115 candidate records
and a 1.8-fold sampling factor to compensate for filtering.
At Assert, target-name leakage checks exclude queries
containing the normalized reference-target name or a
continuous substring of at least four Chinese characters
shared with that name. Finalize removes exact duplicates
and filters near duplicates using character-level 3-gram
Jaccard similarity with a threshold of 0.8. Records retain
the query, spatiotemporal and optional user context, POI
roles, constraints and evidence links, constraint relations,
reference target, satisfiability annotation, and synthesis
provenance.

\begin{figure}[t]
    \centering
    \includegraphics[width=\columnwidth]{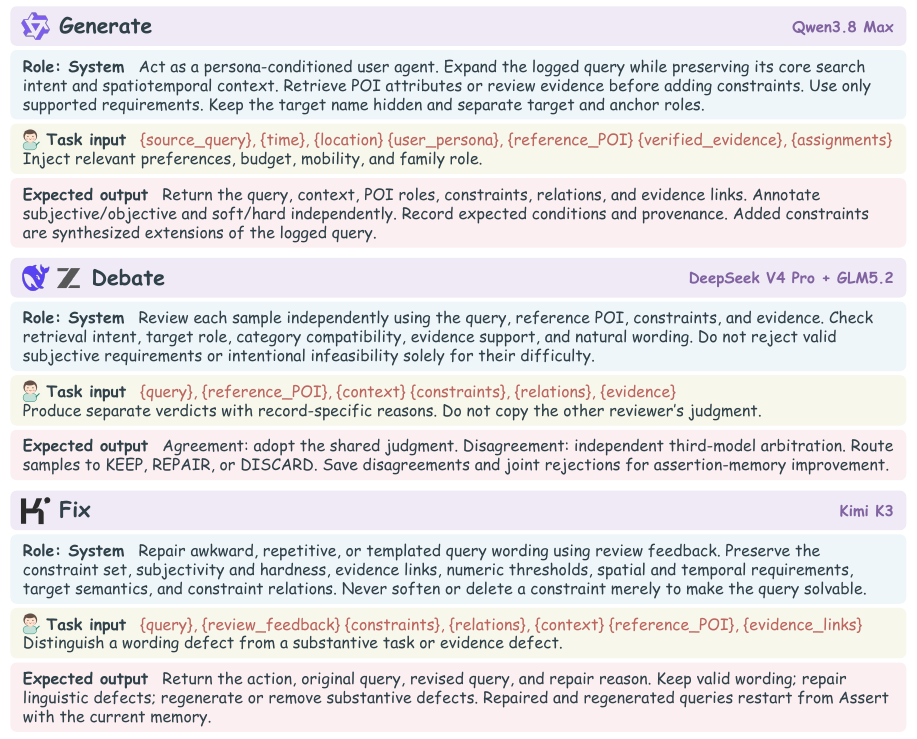
    }
    \caption{Condensed prompts for Generate, Debate, and Fix, showing
instructions, runtime inputs, and expected outputs.
}
    \label{fig:synthesis-prompts}
    \vspace{-10pt}
\end{figure}

Figure~\ref{fig:synthesis-prompts} presents condensed English
adaptations of the Generate, Debate, and Fix prompts,
covering persona-conditioned synthesis, independent review
with arbitration, and query rewriting that preserves
constraints and evidence links.
\subsection{Regional POI Category Distributions}
\label{sec:regional-poi-categories}

\begin{figure}[!t]
    \centering
    \includegraphics[width=\columnwidth,
                     height=0.70\textheight,
                     keepaspectratio]{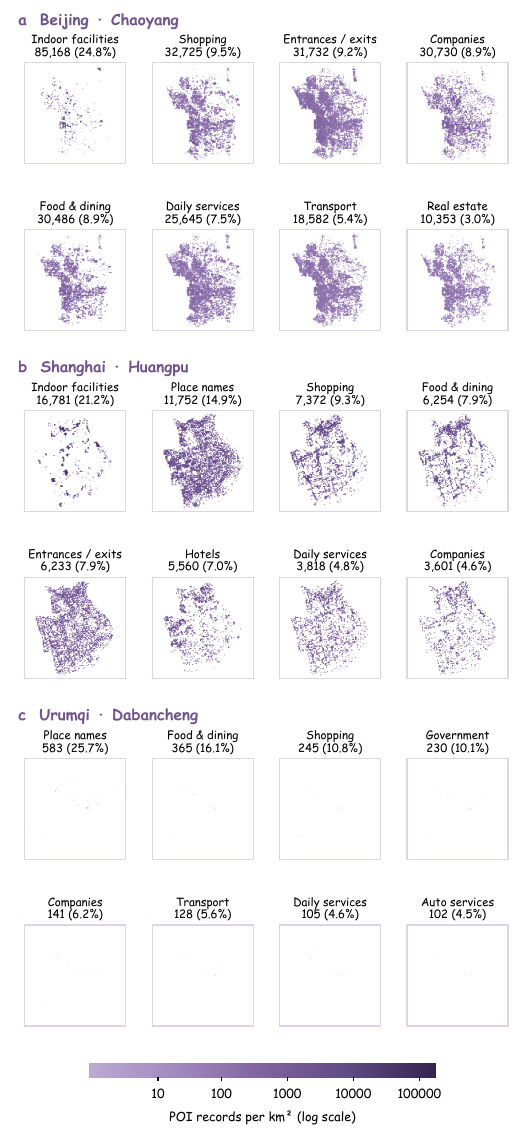}
    \caption{Spatial distributions of the eight most frequent
    POI categories in Chaoyang, Huangpu, and Dabancheng.
    Each panel reports a category's POI count and share of
    the regional collection.}
    \Description{Maps of the leading eight POI categories
    in each of three benchmark regions, with counts,
    percentages, and a shared logarithmic density legend.}
    \label{fig:poi-tag}
    \vspace{-10pt}
\end{figure}

The source POI collections differ in category composition
and spatial concentration. Figure~\ref{fig:poi-tag}
shows the eight most frequent POI categories in each
region, complementing the aggregate regional statistics
in Table~\ref{tab:regional_diversity}.

\subsection{Constraint and Query-Type Distributions}
\label{app:constraint-task-distributions}

\begin{figure}[t]
    \centering
    \includegraphics[width=\columnwidth]{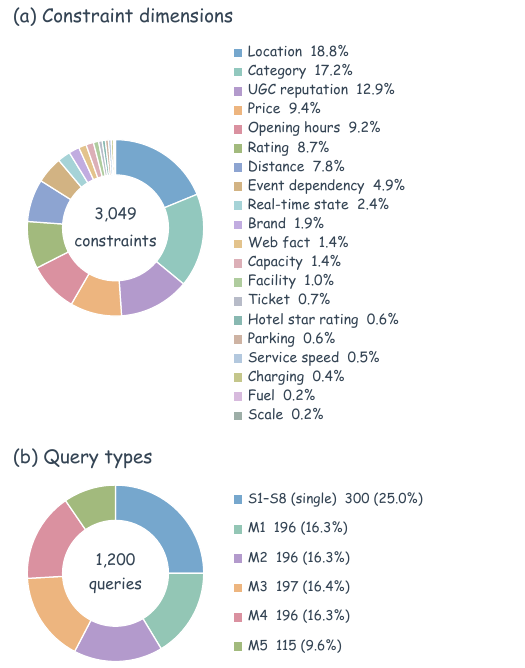}
    \caption{ATLAS-Bench composition. (a) Shares of 3,049 atomic
constraint annotations by dimension. (b) Shares of 1,200 queries by
task type, with S1--S8 aggregated.}

    \Description{Two donut charts showing 20 constraint dimensions
    and six task groups. The largest constraint dimensions are location,
    category, and UGC reputation. Single-constraint queries total 300;
    M1 through M5 contain 196, 196, 197, 196, and 115 queries.}
    \label{fig:benchmark-distribution}
    \vspace{-10pt}
\end{figure}

Figure~\ref{fig:benchmark-distribution} summarizes 3,049 atomic
constraint annotations across 20 observed dimensions and 1,200
queries. Location, category, and UGC reputation account for 48.9\%
of annotations. Single-constraint queries (S1--S8) comprise 25\%
of the benchmark (100 per region), spanning hard/soft and
attribute/review-verified requirements under direct or implicit
phrasing. Multi-constraint queries (M1--M5, 75\%) cover five
regimes: compatible multi-dimensional constraints (\textbf{M1}),
hard--soft tension (\textbf{M2}), soft--soft trade-offs (\textbf{M3}),
mutually exclusive hard conflicts that are unsatisfiable by design
(\textbf{M4}), and external-fact-dependent retrieval (\textbf{M5}). On 133 atomic constraints from 50 sampled queries, human review
agrees with the pipeline's labels on 89.47\% of hard--soft and
97.74\% of subjective--objective annotations.

\section{Metric Formulations and Benchmark Comparisons}
\label{sec:metric_definitions}

\subsection{Formal Computation of Evaluation Metrics}

\paragraph{Intent and Constraint Combination Diversity}
We use the mapping platform's standardized primary category tag as the discrete intent label across a unified vocabulary of $K=23$ categories. Similarly, each atomic requirement is mapped to our 20-type constraint ontology, and each query's sorted, deduplicated constraint types form a unique signature across an invariant empirical space of $M=180$ signatures. Let $p_k$ and $p_m$ denote the regional proportions of intent $k$ and signature $m$, respectively. Both diversities are quantified via normalized Shannon entropy:

\begin{equation}
D_{\mathrm{intent}} = -\frac{\sum_{k=1}^{K} p_k \log p_k}{\log K}, \quad
D_{\mathrm{constraint}} = -\frac{\sum_{m=1}^{M} p_m \log p_m}{\log M}.
\end{equation}
Fixing $M$ across regions prevents sparse collections with few combinations from obtaining artificially inflated scores. Higher values in $[0,1]$ indicate richer, more balanced distributions.

\paragraph{Surface Expression Diversity}
We measure lexical and syntactic variety by inverting leave-one-out character-level $\mathrm{SelfBLEU}_4$ with add-one smoothing. Treating each query $q_i$ as the hypothesis against the remaining $N-1$ queries $R_i = \{q_j \mid j \ne i\}$, surface diversity is defined as:
\begin{equation}
D_{\mathrm{surface}} = 1 - \mathrm{SelfBLEU}_4 = 1 - \frac{1}{N}\sum_{i=1}^{N} \mathrm{BLEU}_4(q_i; R_i).
\end{equation}

\paragraph{Semantic Diversity}
We encode each query into an $\ell_2$-normalized 1,024-dimensional vector $e_i$ using Qwen3-Embedding-0.6B under identical pooling and normalization protocols. Semantic diversity is the mean pairwise cosine distance within a region:
\begin{equation}
D_{\mathrm{semantic}} = \frac{2}{N(N-1)} \sum_{i < j} (1 - e_i^\top e_j).
\end{equation}

\paragraph{Human Authenticity Validation}
We draw a stratified random sample of 200 queries across regions for double-blind inspection. Two annotators independently judge whether each query reflects a plausible real-world user request ($y_i^{(1)}, y_i^{(2)} \in \{0,1\}$). For $N_r$ regional pairs, inter-annotator agreement is $A_r = \frac{1}{N_r} \sum_{i=1}^{N_r} \mathbb{I}(y_i^{(1)} = y_i^{(2)})$, reported alongside the overall proportion of judgments affirming authentic intent. All human annotators participated voluntarily with informed consent.

\subsection{Evaluation Setup Against LocalSearchBench}

Table~\ref{tab:localsearch_comparison} compares ATLAS-Bench with LocalSearchBench across two label-free metrics: semantic diversity via Qwen3-Embedding-0.6B mean pairwise cosine distance, and surface diversity via character level inverted Self-BLEU-4 with add one smoothing. Both measures assign higher values to greater diversity. To ensure comparability, subsets are calibrated to 400 instances through repeated stratified sampling. For LocalSearchBench, we report user questions, generated hop queries, and deduplicated hop queries.

\begin{table}[!htbp]
  \centering
  \caption{Comparison of semantic and surface diversity
  at 400 instances.}
  \label{tab:localsearch_comparison}

  \begingroup
  \scriptsize
  \setlength{\tabcolsep}{3pt}
  \renewcommand{\arraystretch}{1.18}
  \arrayrulecolor{AtlasDiversityGreen!50}

  \begin{tabular*}{\columnwidth}{
    @{\extracolsep{\fill}}lcc@{}}
    \toprule
    \rowcolor{AtlasDiversityGreen!12}
    \textbf{Benchmark View}
      & \textbf{Semantic Dist} $\uparrow$
      & \textbf{Surface Div} $\uparrow$ \\
    \midrule

    \rowcolor{AtlasDiversityGreen!6}
    \textbf{ATLAS-Bench Pooled}
      & 0.5844 & \textbf{0.4622} \\
    ATLAS-Bench Beijing
      & 0.5898 & 0.4067 \\
    ATLAS-Bench Shanghai
      & 0.5630 & 0.4327 \\
    ATLAS-Bench Urumqi
      & 0.5338 & 0.2854 \\

    \addlinespace[3pt]
    LocalSearchBench Question
      & 0.4946 & 0.0629 \\
    LocalSearchBench Hop
      & 0.5830 & 0.1697 \\
    LocalSearchBench Dedup Hop
      & \textbf{0.6028} & 0.1926 \\

    \bottomrule
  \end{tabular*}
  \endgroup
  
\end{table}

\section{Evaluation Metrics}
\label{app:answer-quality-metrics}

We detail the computation of the five answer-quality metrics reported
in Table~\ref{tab:overall-performance}. 
Although valid answers are not restricted to the logged POI, reference-target
hit rate on satisfiable queries correlates strongly with Success@k at Spearman
$\rho=0.93$, confirming that our open-ended constraint evaluation aligns
closely with real user selections.

\noindent\textbf{Notation.}
Let $\mathcal{Q}$ denote all $1{,}200$ queries and
$\mathcal{S} \subset \mathcal{Q}$ the $1{,}004$ queries labeled
satisfiable, excluding the $196$ unsatisfiable queries.
For each query $q$, let $R_q$ be the returned recommendation list,
with $0 \leq |R_q| \leq 10$, and let $n_q$ be the number of annotated
hard constraints and soft preferences.
For recommendation $r \in R_q$, let $m_{qr}$ count the distinct
constraints fully satisfied by evidence belonging to the recommended
POI. Disclosing an unmet or unverifiable soft preference does not
count as satisfying it.
We define $g_{qr}=1$ if $r$ is qualified or acceptable, and
$g_{qr}=0$ otherwise.

\noindent\textbf{CSR and CR.}
Both metrics are micro-averaged over constraint instances across
all returned recommendations.
Let $D=\sum_{q \in \mathcal{Q}} |R_q|n_q$ be their shared denominator:
\[
\begin{aligned}
\mathrm{CSR}
&= \frac{100}{D}
   \sum_{q \in \mathcal{Q}} \sum_{r \in R_q}
   g_{qr}m_{qr}, \quad
\mathrm{CR}
= \frac{100}{D}
   \sum_{q \in \mathcal{Q}} \sum_{r \in R_q}
   m_{qr}.
\end{aligned}
\]
CSR credits satisfied constraints only in qualified or acceptable
recommendations, whereas CR also credits those in unacceptable
recommendations. Thus, $\mathrm{CSR} \leq \mathrm{CR}$.
Empty answers contribute neither to the numerators nor to the
shared denominator.

\noindent\textbf{Success@k and Recs/query.}
These metrics are averaged over the fixed should-answer set
$\mathcal{S}$:
\[
\begin{aligned}
\mathrm{Success@}k
&= \frac{100}{|\mathcal{S}|}
   \sum_{q \in \mathcal{S}}
   \mathbf{1}\!\left[
     \sum_{r \in R_q} g_{qr} > 0
   \right], \\
\mathrm{Recs/query}
&= \frac{1}{|\mathcal{S}|}
   \sum_{q \in \mathcal{S}} |R_q|.
\end{aligned}
\]
Success@k evaluates the full returned list, containing at most
$10$ recommendations, and requires at least one qualified or
acceptable recommendation.
Recs/query counts all returned recommendations, regardless of
their verdicts.
Empty answers contribute zero to both metrics while remaining
in the denominator, so unanswered satisfiable queries are
explicitly accounted for.

\noindent\textbf{Factual accuracy.}
For each query $q$, let $b_q$ and $f_q$ be binary indicators for
passing the deterministic grounding check and the LLM factuality
check, respectively.
The grounding check requires the answer's
\texttt{no\_exact\_match} flag to be consistent with its returned
recommendations, and every cited evidence ID to exist in the
observed tool results and belong to the recommended POI.
The DeepSeek-V4-Pro factuality judge sets $f_q=1$ only when the
answer contains no factual errors or unsupported claims relative
to the observed evidence. This metric is computed over all $1{,}200$ queries.
An answer receives credit only if both checks pass:
\[
\mathrm{Factual\ accuracy}
= \frac{100}{|\mathcal{Q}|}
  \sum_{q \in \mathcal{Q}} b_q f_q.
\]

\noindent\textbf{Trajectory quality and efficiency.}
Each trajectory dimension is scored out of $5$ and averaged over
$\mathcal{Q}$. Efficiency metrics directly average the per-query
execution statistics logged by ATLASBox.

\section{Revisions to the Assertion Layer}
\label{app:assertion-evolution}

Table~\ref{tab:assertion-revisions} summarizes anonymized
updates to the assertion memory and validation logic.
These updates document implementation changes, rather
than verified improvements to all released samples.

\begin{table}[!htbp]
  \centering
  \caption{Representative assertion and validation revisions.}
  \label{tab:assertion-revisions}

  \begingroup
  \scriptsize
  \setlength{\tabcolsep}{3pt}
  \renewcommand{\arraystretch}{1.08}
  \arrayrulecolor{AtlasRevisionPurple!50}

  \begin{tabular}{@{}
    >{\raggedright\arraybackslash}p{0.21\columnwidth}
    >{\raggedright\arraybackslash}p{0.31\columnwidth}
    >{\raggedright\arraybackslash}
      p{\dimexpr0.48\columnwidth-12pt\relax}
    @{}}
    \toprule
    \rowcolor{AtlasRevisionPurple!12}
    \textbf{Check}
      & \textbf{Observed issue}
      & \textbf{Revision} \\
    \midrule

    Name leakage
      & Geographic anchors falsely flagged.
      & Exempt geographic terms; fix whitespace matching. \\
    \addlinespace[2pt]

    \rowcolor{AtlasRevisionPurple!4}
    Evidence linkage
      & POI-based joins mixed search events.
      & Join evidence by source-record ID. \\
    \addlinespace[2pt]

    Semantic rejection
      & Lexical rules rejected valid queries.
      & Broaden matching; make the check advisory. \\
    \addlinespace[2pt]

    \rowcolor{AtlasRevisionPurple!4}
    Conflict references
      & Mixed indexing linked wrong constraints.
      & Normalize index conventions. \\
    \addlinespace[2pt]

    Constraint types
      & Overlapping fields conflated dimensions.
      & Refine matching; separate semantic dimensions. \\
    \addlinespace[2pt]

    \rowcolor{AtlasRevisionPurple!4}
    Satisfiability
      & Default labels masked conflicts.
      & Remove positive defaults; repair conflict logic. \\
    \addlinespace[2pt]

    Check scope
      & Conflict checks affected unrelated cases.
      & Restrict checks to relevant conflict decisions. \\
    \addlinespace[2pt]

    \rowcolor{AtlasRevisionPurple!4}
    Input coverage
      & Missing shards and unrelated files.
      & Load all shards using explicit file patterns. \\
    \addlinespace[2pt]

    Consistency
      & Duplicate checks and fixed vocabularies drifted.
      & Share detectors; configure regional vocabularies. \\
    \addlinespace[2pt]

    \rowcolor{AtlasRevisionPurple!4}
    Operating status
      & Closure evidence was missed.
      & Check both source name fields. \\

    \bottomrule
  \end{tabular}
  \endgroup
\end{table}

\section{Available Tools}
\label{app:tools}
Table~\ref{tab:tools} summarizes the tools available
in ATLASBox. Returned records include evidence
identifiers for citation in the final answer.

\begin{table}[!htbp]
  \centering
  \caption{ATLASBox search tools and required inputs.
  Optional arguments control result limits and filters.}
  \label{tab:tools}

  \begingroup
  \footnotesize
  \setlength{\tabcolsep}{4pt}
  \renewcommand{\arraystretch}{1.3}
  \renewcommand{\tabularxcolumn}[1]{m{#1}}
  \arrayrulecolor{AtlasToolBlue!55}

  \begin{tabularx}{\columnwidth}{@{}
    >{\raggedright\arraybackslash
      \color{AtlasToolBlue}\ttfamily}m{0.28\columnwidth}
    >{\raggedright\arraybackslash}X
    >{\raggedright\arraybackslash
      \ttfamily}m{0.22\columnwidth}@{}}
    \toprule
    \rowcolor{AtlasToolBlue!12}
    \normalfont\bfseries Tool
      & \textbf{Purpose}
      & \normalfont\bfseries Inputs \\
    \midrule

    search\_pois
      & Retrieve POIs and structured attributes;
        resolve landmark coordinates.
      & query \\

    \rowcolor{AtlasToolBlue!4}
    search\_notes
      & Retrieve review evidence for experiential requirements.
      & query \\

    get\_poi\_notes
      & Inspect reviews linked to a specific POI.
      & poiid \\

    \rowcolor{AtlasToolBlue!4}
    search\_nearby
      & Find POIs within a radius of the given coordinates.
      & query, loc,\newline distance\_m \\

    \bottomrule
  \end{tabularx}
  \endgroup
\end{table}

\end{document}